\documentclass{sigkddExp}

\usepackage{booktabs}            
\usepackage{multirow}            
\usepackage{amsfonts}            
\usepackage{graphicx}            
\usepackage{duckuments}          
\usepackage{xcolor}

\graphicspath{{./images/}}

\usepackage{multirow}
\usepackage{multicol}
\usepackage{caption}
\usepackage{subcaption}
\usepackage{bm}
\usepackage{color}
\usepackage{adjustbox}

\usepackage{enumitem}
\usepackage{bbm}
\usepackage{graphicx}
\usepackage{mathtools}
\usepackage{bbm}
\usepackage{balance}
\usepackage{graphics}
\usepackage{bbding}
\usepackage{algorithm}
\usepackage{algpseudocode}
\usepackage{booktabs} 
\usepackage{tcolorbox}
\usepackage{colortbl}
\usepackage{amsmath,amsfonts}
\usepackage{booktabs}
\usepackage{multirow}
\usepackage{xfrac}
\usepackage{tabu}
\usepackage{xcolor}
\usepackage{multirow}
\usepackage{array}

\usepackage{url}
\usepackage{paralist}
\usepackage{color}
\usepackage{graphicx}
\usepackage{xspace}

\newcommand{\cora}{\textsc{Cora}\xspace}
\newcommand{\citeseer}{\textsc{Citeseer}\xspace}
\newcommand{\pubmed}{\textsc{Pubmed}\xspace}
\newcommand{\arxiv}{\textsc{Ogbn-arxiv}\xspace}
\newcommand{\products}{\textsc{Ogbn-products}\xspace}

\newcommand{\fen}{\textit{\textbf{feature-level enhancement}}}
\newcommand{\ten}{\textit{\textbf{text-level enhancement}}}

\usepackage{tabularx}

\newcommand{\prompt}[1]{\textbf{$\langle$#1$\rangle$}}

\usepackage{perpage} 
\MakePerPage{footnote} 
\usepackage{float}

\usepackage[normalem]{ulem}
\useunder{\uline}{\ul}{}

\begin{document}

%

\title{Exploring the Potential of Large Language Models (LLMs) \\in Learning on Graphs}

\author{
Zhikai Chen$^{1}$, Haitao Mao$^1$, Hang Li$^1$, Wei Jin$^3$, Hongzhi Wen$^1$, \\ 
Xiaochi Wei$^2$, Shuaiqiang Wang$^2$, Dawei Yin$^2$ \\
Wenqi Fan$^4$, Hui Liu$^1$, Jiliang Tang$^1$ \\
$^1$Michigan State University  \quad  $^2$ Baidu Inc.  \quad $^3$ Emory University \\ $^4$ The Hong Kong Polytechnic University \\
\{chenzh85, haitaoma, lihang4, wenhongz, liuhui7, tangjili\}@msu.edu, \\
\{weixiaochi, wangshuaiqiang\}@baidu.com, yindawei@acm.org,\\
wei.jin@emory.edu, \\ 
wenqifan03@gmail.com \\
}

\maketitle

\begin{abstract}
Learning on Graphs has attracted immense attention due to its wide real-world applications. The most popular pipeline for learning on graphs with textual node attributes primarily relies on Graph Neural Networks (GNNs), and utilizes shallow text embedding as initial node representations, which has limitations in general knowledge and profound semantic understanding. In recent years, Large Language Models (LLMs) have been proven to possess extensive common knowledge and powerful semantic comprehension abilities that have revolutionized existing workflows to handle text data. In this paper, we aim to explore the potential of LLMs in graph machine learning, especially the node classification task, and investigate two possible pipelines: \textit{LLMs-as-Enhancers} and \textit{LLMs-as-Predictors}. The former leverages LLMs to enhance nodes' text attributes with their massive knowledge and then generate predictions through GNNs. The latter attempts to directly employ LLMs as standalone predictors. We conduct comprehensive and systematical studies on these two pipelines under various settings. From comprehensive empirical results, we make original observations and find new insights that open new possibilities and suggest promising directions to leverage LLMs for learning on graphs. Our codes and datasets are available at: \url{https://github.com/CurryTang/Graph-LLM}
\end{abstract}.

\section{Introduction}
\label{intro}


\vspace{1em}
Graphs are ubiquitous in various disciplines 
and applications, encompassing a wide range of 
real-world scenarios~\cite{Xia2021GraphLA}. Many of 
these graphs have nodes that are associated 
with text attributes, resulting in the 
emergence of text-attributed graphs, such as citation graphs~\cite{hu2020open, Sen_Namata_Bilgic_Getoor_Galligher_Eliassi-Rad_2008} and product graphs~\cite{Chiang2019ClusterGCNAE}. For example, in the \products dataset~\cite{hu2020open}, each node 
represents a product, and its corresponding 
textual description is treated as the node's attribute. 
These graphs have seen widespread use across 
a myriad of domains, from social network analysis~\cite{social_network}, information retrieval~\cite{Zhu2021TextGNNIT}, 
to a diverse range of natural language processing tasks~\cite{liu-etal-2020-fine, Yao2018GraphCN}. 

Given the prevalence of text-attributed graphs (TAGs), we aim to explore how to effectively handle these graphs, with a focus on the node classification task. Intuitively, TAGs provide both node attribute and graph structural information. Thus, it is important to effectively capture both while modeling their interrelated correlation. Graph Neural Networks (GNNs)~\cite{ma2021deep} have emerged as the de facto technique for handling graph-structured data, often leveraging a message-passing paradigm to effectively capture the graph structure. To encode textual information, conventional pipelines typically make use of non-contextualized shallow embeddings e.g., Bag-of-Words~\cite{harris1954distributional} and Word2Vec~\cite{mikolov2013efficient} embeddings, as seen in the common graph benchmark datasets~\cite{hu2020open, Sen_Namata_Bilgic_Getoor_Galligher_Eliassi-Rad_2008}, where GNNs are subsequently employed to process these embeddings. Recent studies demonstrate that these non-contextualized shallow embeddings suffer from some limitations, such as the inability to capture polysemous words~\cite{Qiu2020PretrainedMF} and deficiency in semantic information~\cite{miaschi-dellorletta-2020-contextual, ethayarajh-2019-contextual}, which may lead to sub-optimal performance on downstream tasks.

Compared to these non-contextualized shallow textual embeddings, large language models (LLMs) present massive context-aware knowledge and superior semantic comprehension capability through the process of pre-training on large-scale text corpora~\cite{Petroni2019LanguageMA, ethayarajh-2019-contextual}. 
This knowledge achieved from pre-training has led to a surge of revolutions for downstream NLP tasks~\cite{Zhao2023ASO}.
Exemplars such as ChatGPT and GPT4~\cite{OpenAI2023GPT4TR}, equipped with hundreds of billions of parameters, exhibit superior performance~\cite{Bubeck2023SparksOA} on numerous text-related tasks from various domains. Considering the exceptional ability of these LLMs to process and understand textual data, a pertinent question arises: (1) \textit{Can we leverage the knowledge of LLMs to compensate for the deficiency of contextualized knowledge and semantic comprehension inherent in the conventional GNN pipelines?}
In addition to the knowledge learned via pre-training, recent studies suggest that LLMs present preliminary success on tasks with implicit graph structures such as recommendation \cite{liu2023chatgpt_rec, Gao2023ChatRECTI}, ranking \cite{Ji2023ExploringCA}, and multi-hop reasoning \cite{creswell2023selectioninference}, in which LLMs are adopted to make the final predictions. Given such success, we further question: (2) \textit{Can LLMs, beyond merely integrating with GNNs, independently perform predictive tasks with explicit graph structures? } In this paper, we aim to embark upon a preliminary investigation of these two  questions by undertaking a series of extensive empirical analyses. Particularly, the key challenge is how to design an LLM-compatible pipeline for graph learning tasks. Consequently, we explore two potential pipelines to incorporate LLMs: (1) \textit{LLMs-as-Enhancers}: LLMs are adopted to enhance the textual information; subsequently, GNNs utilize refined textual data to generate predictions. (2) \textit{LLMs-as-Predictors}: LLMs are adapted to generate the final predictions, where structural and attribute information is present completely through natural languages.

In this work, we embrace the challenges and opportunities to study the utilization of LLMs in graph-related problems and aim to deepen our understanding of \textit{the potential of LLMs on graph machine learning}, with a focus on the node classification task. \textbf{First}, we aim to investigate how LLMs can enhance GNNs by leveraging their extensive knowledge and semantic comprehension capability. It is evident that different types of LLMs possess varying levels of capability, and more powerful models often come with more usage restrictions~\cite{sun2022black, Zhao2023ASO, Qiu2020PretrainedMF}. Therefore, we strive to design different strategies tailored to different types of models, and better leverage their capabilities within the constraints of these usage limitations. \textbf{Second}, we want to explore how LLMs can be adapted to explicit graph structures as a predictor. A principal challenge lies in crafting a prompt that enables the LLMs to effectively use structural and attribute information. To address this challenge, we attempt to explore what information can assist LLMs in better understanding and utilizing graph structures. Through these investigations, we make some insightful observations and gain a better understanding of the capabilities of LLMs in graph machine learning.

\textbf{Contributions.} Our contributions are summarized as follows:
\begin{compactenum}[1.]
\item We explore two pipelines that incorporate LLMs to handle TAGs: \textit{LLMs-as-Enhancers} and \textit{LLMs-as-Predictors}. The first pipeline treats the LLMs as attribute enhancers, seamlessly integrating them with GNNs. The second pipeline directly employs the LLMs to generate predictions.

\item For \textit{LLMs-as-Enhancers}, we introduce two strategies to enhance text attributes via LLMs. We further conduct a series of experiments to compare the effectiveness of these enhancements. 
\item For \textit{LLMs-as-Predictors}, we design a series of experiments to explore LLMs' capability in utilizing structural and attribute information. From empirical results, we summarize some original observations and provide new insights.  

\end{compactenum}


\textbf{Key Insights.} Through comprehensive empirical evaluations, we find the following key insights: 
\begin{compactenum}[1.]
\item For \textit{LLMs-as-Enhancers}, using deep sentence embedding 
models to generate embeddings for node 
attributes show both effectiveness and efficiency.

\item  For \textit{LLMs-as-Enhancers}, utilizing LLMs to augment node attributes at the text level also leads to improvements in downstream performance.

\item  For \textit{LLMs-as-Predictors}, LLMs present preliminary effectiveness but we should be careful about their inaccurate predictions and the potential test data leakage problem. 

\item LLMs demonstrate the potential to serve as good annotators for labeling nodes, as a decent portion of their annotations is accurate.  
\end{compactenum}

\textbf{Organization.} The remaining of this paper is organized as follows.  Section~\ref{sec:prl} introduces necessary preliminary knowledge and notations used in this paper. Section~\ref{sec: pipeline} introduces two pipelines to leverage LLMs under the task of node classification.  Section~\ref{sec:enh} explores the first pipeline, \textit{LLMs-as-Enhancers}, which adopts LLMs to enhance text attributes. Section~\ref{sec:pred} details the second pipeline, \textit{LLMs-as-Predictors}, exploring the potential for directly applying LLMs to solve graph learning problems as a predictor. Section~\ref{sec:rw} discusses works relevant to the applications of LLMs in the graph domain.  Section~\ref{sec:fut} summarizes our insights and discusses the limitations of our study and the potential directions of LLMs in the graph domain. 


\section{Preliminaries}
\label{sec:prl}
\vspace{1em}
In this section, we present concepts, notations and problem settings used in the work. We primarily delve into the node classification task on the text-attributed graphs, which is one of the most important downstream tasks in the graph learning domain. Next, we first give the definition of text-attributed graphs. 


\noindent{}\textbf{Text-Attributed Graphs}
A text-attributed graph (TAG) $\mathcal{G}_{S}$ is defined as a structure consisting of nodes $\mathcal{V}$ and their corresponding adjacency matrix $\mathbf{A} \in \mathbb{R}^{|V|\times |V|}$. For each node $v_{i} \in \mathcal{V}$, it is associated with a text attribute, denoted as $\mathbf{s}_i$.

In this study, we focus on node classification, which is one of the most commonly adopted graph-related tasks. 

\noindent{}\textbf{Node Classification on TAGs}
Given a set of labeled nodes $\mathcal{L} \subset \mathcal{V}$ with their labels $y_{\mathcal{L}}$, we aim to predict the labels $\mathbf{y}_\mathcal{U}$ for the remaining unlabeled nodes $\mathcal{U} = \mathcal{V} \: \backslash \: \mathcal{L}$.

We use the citation network dataset \arxiv~\cite{hu2020open} as an illustrative example. In such a graph, each node represents an individual paper from the computer science subcategory, with the attribute of the node embodying the paper's title and abstracts. The edges denote the citation relationships. The task is to classify the papers into their corresponding categories, for example, ``cs.cv'' (i.e., computer vision). Next, we introduce the models adopted in this study, including graph neural networks and large language models. 

\noindent{}\textbf{Graph Neural Networks.}
    When applied to TAGs for node classification, Graph Neural Networks (GNNs) leverage the structural interactions between nodes. Given initial node features $h_{i}^{0}$, GNNs update the representation of each node by aggregating the information from neighboring nodes in a message-passing manner~\cite{Gilmer2017NeuralMP}. The $l$-th layer can be formulated as:
\begin{equation}
    h_i^{l}=\operatorname{UPD}^{l}\left(h_i^{l-1}, \operatorname{AGG}_{j \in \mathcal{N}(i)} \operatorname{MSG}^{l}\left(h_i^{l-1}, h_j^{l-1}\right)\right), 
\end{equation}
where $\operatorname{AGG}$ is often an aggregation function such as summation, or maximum. $\operatorname{UPD}$ and $\operatorname{MSG}$ are usually some differentiable functions, such as MLP.  The final hidden representations can be passed through a fully connected layer to make classification predictions. 

\noindent\textbf{Large Language Models.}
In this work, we primarily utilize the term "large language models" (LLMs) to denote language models that have been pre-trained on extensive text corpora. Despite the diversity of pre-training objectives~\cite{bert, radford2019language, 2020t5}, the shared goal of these LLMs is to harness the knowledge acquired during the pre-training phase and repurpose it for a range of downstream tasks. Based on their interfaces, specifically considering whether their embeddings are accessible to users or not, in this work we roughly classify LLMs as below:

\noindent{}\textbf{Embedding-visible LLMs}
    Embedding-visible LLMs provide access to their embeddings, allowing users to interact with and manipulate the underlying language representations. Embedding-visible LLMs enable users to extract embeddings for specific words, sentences, or documents, and perform various natural language processing tasks using those embeddings. Examples of embedding-visible LLMs include BERT~\cite{bert}, Sentence-BERT~\cite{sbert}, and Deberta~\cite{he2020deberta}.

\noindent{}\textbf{Embedding-invisible LLMs}
    Embedding-invisible LLMs do not provide direct access to their embeddings or allow users to manipulate the underlying language representations. Instead, they are typically deployed as web services~\cite{sun2022black} and offer restricted interfaces. For instance, ChatGPT~\cite{OpenAI2022}, along with its API, solely provides a text-based interface. Users can only engage with these LLMs through text interactions.

In addition to the interfaces, the size, capability, and model structure are crucial factors in determining how LLMs can be leveraged for graphs. Consequently, we take into account the following four types of LLMs: 

\begin{compactenum}[1.]
    \item \textbf{\textit{Pre-trained Language Models:}} We use the term "pre-trained language models" (PLMs) to refer to those relatively small large language models, such as Bert~\cite{bert} and Deberta~\cite{he2020deberta}, which can be fine-tuned for downstream tasks. It should be noted that strictly speaking, all LLMs can be viewed as PLMs. Here we adopt the commonly used terminology for models like BERT~\cite{Qiu2020PretrainedMF} to distinguish them from other LLMs follwing the convention in a recent paper~\cite{Zhao2023ASO}. 
    \item \textbf{\textit{Deep Sentence Embedding Models:}} These models typically use PLMs as the base encoders and adopt the bi-encoder structure~\cite{sbert, wang2022text, Neelakantan2022TextAC}. They further pre-train the models in a supervised~\cite{sbert} or contrastive manner~\cite{wang2022text, Neelakantan2022TextAC}. In most cases, there is no need for these models to conduct additional fine-tuning for downstream tasks. These models can be further categorized into \textit{local sentence embedding models} and \textit{online sentence embedding models}. \textit{Local sentence embedding models} are open-source and can be accessed locally, with Sentence-BERT (SBERT) being an example. On the other hand, \textit{online sentence embedding models} are closed-source and deployed as services, with OpenAI's text-ada-embedding-002~\cite{Neelakantan2022TextAC} being an example. 
    \item \textbf{\textit{Large Language Models:}} Compared to PLMs, Large Language Models (LLMs) exhibit significantly enhanced capabilities with orders of magnitude more parameters. LLMs can be categorized into two types. The first type consists of open-source LLMs, which can be deployed locally, providing users with transparent access to the models' parameters and embeddings. However, the substantial size of these models poses a challenge, as fine-tuning them can be quite cumbersome. One representative example of an open-source LLM is LLaMA~\cite{llama}.
    The second type of LLMs is typically deployed as services~\cite{sun2022black}, with  restrictions placed on user interfaces. In this case, users are unable to access the  model parameters, embeddings, or logits directly.  The most powerful LLMs such as  ChatGPT~\cite{OpenAI2022} and GPT4~\cite{OpenAI2023GPT4TR} belong to this kind.

\end{compactenum}

Among the four types of LLMs,  PLMs, deep sentence embedding models, and open-source LLMs are often embedding-visible LLMs. Closed-source LLMs are embedding-invisible LLMs.

\section{Pipelines for LLMs in Graphs} 
\label{sec: pipeline}
\vspace{1em}

\begin{figure*}[!htb]
    \centering
    \begin{subfigure}[b]{0.48\textwidth}
        \centering
        \includegraphics[width=0.8\textwidth]{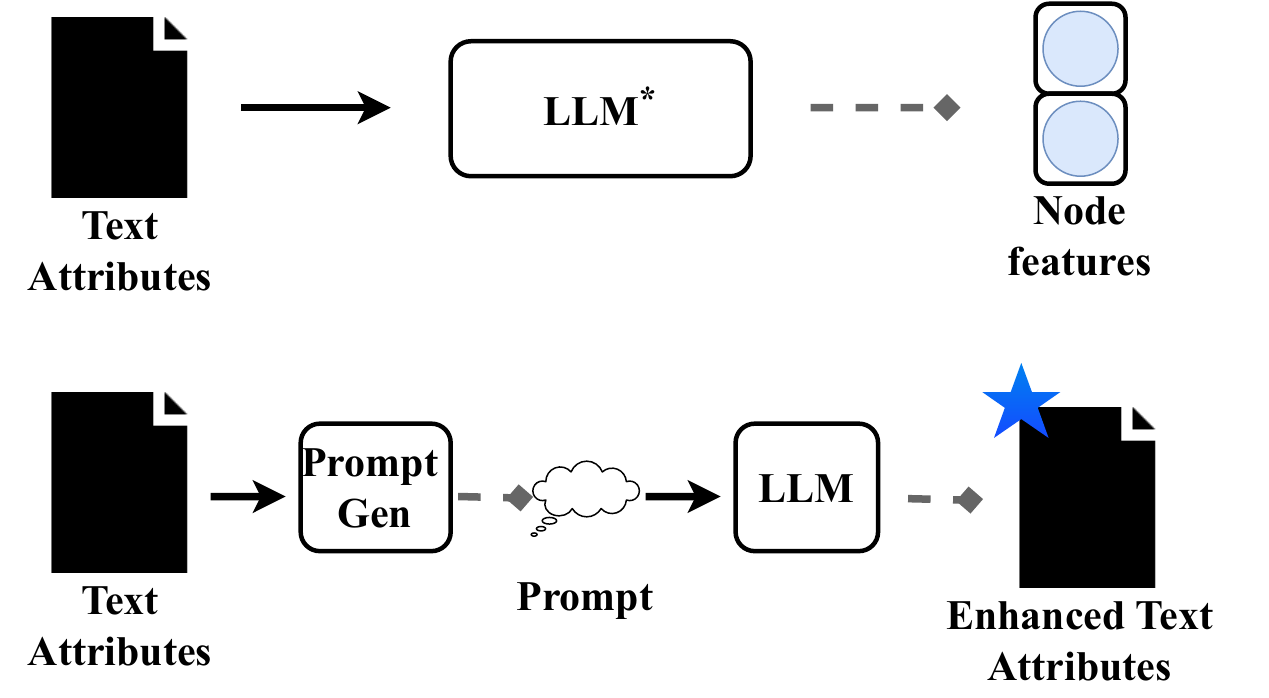}
        \caption{\textit{An illustration of LLMs-as-Enhancers}, where LLMs pre-process the text attributes, and GNNs eventually make the predictions. Three different structures for this pipeline are demonstrated in Figure~\ref{llm_pipeline}.}
        \label{llm_enhancer}
    \end{subfigure}
    \hfill
    \begin{subfigure}[b]{0.50\textwidth}
        \centering
        \includegraphics[width=\textwidth]{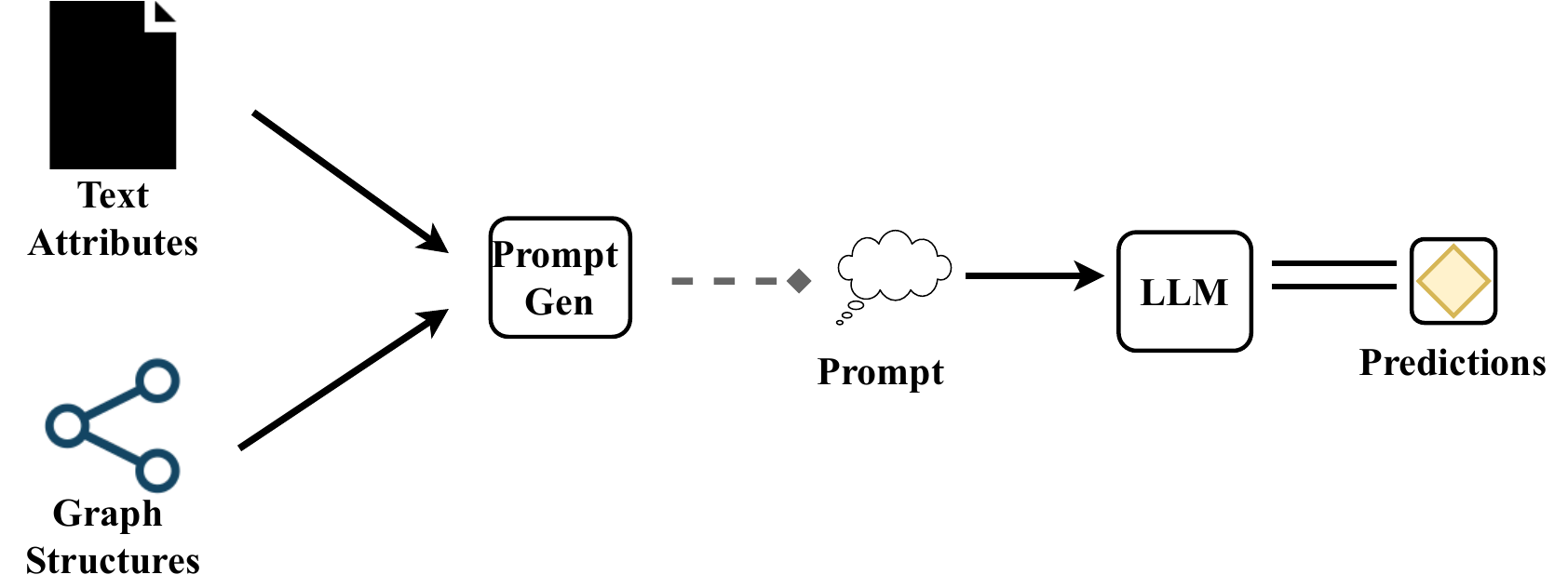}
        \caption{\textit{An illustration of LLMs-as-Predictors}, where LLMs directly make the predictions. The key component for this pipeline is how to design an effective prompt to incorporate structural and attribute information. }
        \label{llm_single}
    \end{subfigure}
    
    \caption{Pipelines for integrating LLMs into graph learning. In all figures, we use ``PLM'' to denote small-scale PLMs that can be fine-tuned on downstream datasets,  ``LLM\textsuperscript{*}'' to denote embedding-visible LLMs, and ``LLM'' to denote embedding-invisible LLMs. } 
    \label{pipeline}
\end{figure*}

\begin{figure*}[h]
    \centering
    \includegraphics[width=0.8\textwidth]{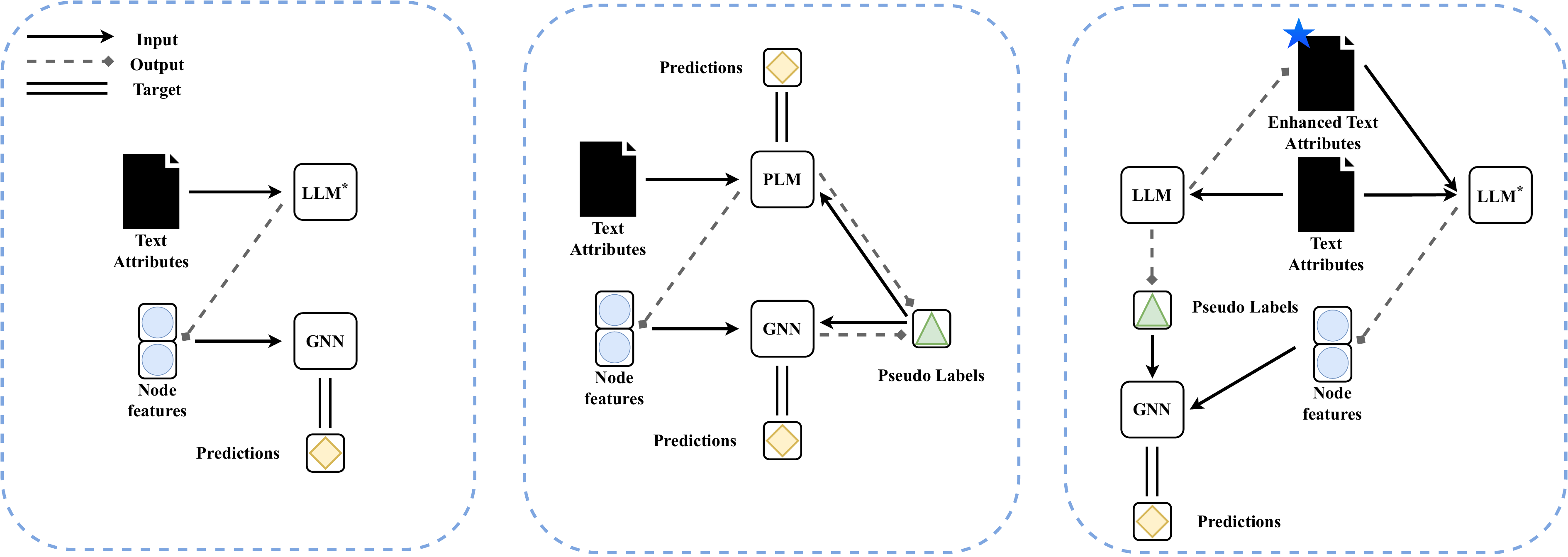}
    \caption{Three strategies to adopt LLMs as enhancers. The first two integrating structures are designed for feature-level enhancement, while the last structure is designed for text-level enhancement. From left to right: (1) Cascading Structure: Embedding-visible LLMs enhance text attributes directly by encoding them into initial node features for GNNs. (2) Iterative Structure: GNNs and PLMs are co-trained in an iterative manner. (3) Text-level enhancement structure: Embedding-invisible LLMs are initially adopted to enhance the text attributes by generating augmented attributes. The augmented attributes and original attributes are encoded and then ensembled together. } 
\label{llm_pipeline}
\end{figure*}

Given the superior power of LLMs in understanding textual information, we now investigate different strategies to leverage LLMs for node classification in textual graphs. Specifically,  we present two distinct pipelines:  \textit{LLMs-as-Enhancers} and \textit{LLMs-as-Predictors}. Figure~\ref{pipeline} provides figurative illustrations of these two pipelines, and we elaborate on their details as follows.

\noindent{}{\textbf{\textit{LLMs-as-Enhancers}}} In this pipeline, LLMs are leveraged to enhance the text attributes. As shown in Figure~\ref{pipeline}, for \textit{LLMs-as-Enhancers}, LLMs are adopted to pre-process the text attributes, and then GNNs are trained on the enhanced attributes as the predictors. Considering different structures of LLMs, we conduct enhancements either at the \textbf{feature level} or at the \textbf{text level} as shown in Figure~\ref{llm_pipeline}. 
\begin{compactenum}[1.]
    \item \textit{\textbf{Feature-level enhancement:}} For feature-level enhancement, embedding-visible LLMs inject their knowledge by simply encoding the text attribute $s_{i}$ into text embeddings $h_{i} \in R^{d}$. We investigate two feasible \textbf{integrating structures} for feature-level enhancement. \textbf{(1) Cascading structure:} Embedding-visible LLMs and GNNs are combined sequentially. Embedding-visible LLMs first encode text attributes into text features, which are then adopted as the initial node features for GNNs. \textbf{(2) Iterative structure~\cite{GLEM}:} PLMs and GNNs are co-trained together by generating pseudo labels for each other. Only PLMs are suitable for this structure since it involves fine-tuning. 
    \item \textit{\textbf{Text-level enhancement:}} For text-level enhancement, given the text attribute $s_{i}$, LLMs will first transform the text attribute into augmented attribute $s_{i}^{Aug}$. Enhanced attributes will then be encoded into enhanced node features $h_{i}^{Aug} \in R^{d}$ through embedding-visible LLMs. GNNs will make predictions by ensembling the original node features and augmented node features.
\end{compactenum}

\noindent{}{\textbf{\textit{LLMs-as-Predictors}}} In this pipeline, LLMs are leveraged to directly make predictions for the node classification task. As shown in Figure~\ref{llm_single}, for \textit{LLMs-as-Predictors}, the first step is to  design prompts to represent graph structural information, text attributes, and label information with texts. Then, embedding-invisible LLMs make predictions based on the information embedded in the prompts.

\section{LLMs as the Enhancers}
\label{sec:enh}

\vspace{1em}
In this section, we investigate the potential of employing LLMs to enrich the text attributes of nodes. As presented in Section~\ref{sec: pipeline}, we consider \textbf{\textit{feature-level enhancement}}, which injects LLMs' knowledge by encoding text attributes into features. Moreover, we consider \textbf{\textit{text-level enhancement}}, which inject LLMs' knowledge by augmenting the text attributes at the text level. We first study \textbf{\textit{feature-level enhancement}}.

\subsection{Feature-level Enhancement} 
\vspace{1.5em}
In \textit{feature-level enhancement}, we mainly study how to combine embedding-visible LLMs with GNNs at the feature level. The embedding generated by LLMs will be adopted as the initial features of GNNs. We first briefly introduce the dataset and dataset split settings we use.

\noindent{}\textbf{Datasets.} In this study, we adopt \cora~\cite{McCallum2000AutomatingTC}, \pubmed~\cite{Sen_Namata_Bilgic_Getoor_Galligher_Eliassi-Rad_2008}, \arxiv, and \products~\cite{hu2020open}, four popular benchmarks for node classification. We present their detailed statistics and descriptions in Appendix~\ref{app: real-world}. Specifically, we examine two classification dataset split settings, specifically tailored for the \cora and \pubmed datasets. Meanwhile, for \arxiv and \products, we adopt the official dataset splits.
(1) For \cora and \pubmed, the first splitting setting addresses \textbf{low-labeling-rate} conditions, which is a commonly adopted setting~\cite{Yang2016RevisitingSL}. To elaborate, we randomly select 20 nodes from each class to form the training set. Then, 500 nodes are chosen for the validation set, while 1000 additional random nodes from the remaining pool are used for the test set.  
(2) The second splitting setting caters to \textbf{high-labeling-rate} scenarios, which is also a commonly used setting, and also adopted by TAPE~\cite{he2023explanations}. In this setting, 60\% of the nodes are designated for the training set, 20\% for the validation set, and the remaining 20\% are set aside for the test set. We take the output of GNNs and compare it with the ground truth of the dataset. We conduct all the experiments on 10 different seeds and report both average accuracy and variance. 

\noindent{}\textbf{Baseline Models.} In our exploration of how LLMs augment node attributes at the feature level, we consider three main components: (1) \textit{Selection of GNNs}, (2) \textit{Selection of LLMs}, and (3) \textit{Intergrating structures for LLMs and GNNs}. In this study, we choose the most representative models for each component, and the details are listed below. 

\begin{compactenum}[1.]
    \item \textit{Selection of GNNs:} For GNNs on \cora and \pubmed, we consider Graph Convolutional Network (GCN)~\cite{kipf2017semisupervised} and Graph Attention Network~(GAT)~\cite{veličković2018graph}. We also include the performance of MLP to \textbf{evaluate the quality of text embeddings without aggregations}. For \arxiv, we consider GCN, MLP, and a better-performed GNN model RevGAT~\cite{li2021training}. For \products, we consider GraphSAGE~\cite{hamilton2017inductive} which supports neighborhood sampling for large graphs, MLP, and a state-of-the-art model SAGN~\cite{sun2021scalable}. For RevGAT and SAGN, we adopt all tricks utilized in the OGB leaderboard~\cite{hu2020open}\footnote{\url{https://ogb.stanford.edu/docs/leader_nodeprop/}}.

    \item \textit{Selection of LLMs:} 
    To enhance the text attributes at the feature level, we specifically require embedding-visible LLMs. Specifically, we select (1) \textbf{Fixed PLM/LLMs without fine-tuning:} We consider Deberta~\cite{he2020deberta} and LLaMA~\cite{llama}. The first one is adapted from  GLEM~\cite{GLEM} and we follow the setting of GLEM~\cite{GLEM} to adopt the [CLS] token of PLMs as the text embeddings. LLaMA is a widely adopted open-source LLM, which has also been included in Langchain\footnote{\url{https://python.langchain.com/}}. We adopt LLaMA-cpp\footnote{\url{https://github.com/ggerganov/llama.cpp}}, which adopt the [EOS] token as text embeddings in our experiments. (2) \textbf{Local sentence embedding models}: We adopt Sentence-BERT ~\cite{sbert} and e5-large~\cite{wang2022text}. The former is one of the most popular lightweight deep text embedding models while the latter is the state-of-the-art model on the MTEB leaderboard~\cite{muennighoff-etal-2023-mteb}. (3) \textbf{Online sentence embedding models}: We consider two online sentence embedding models, i.e., text-ada-embedding-002~\cite{Neelakantan2022TextAC} from OpenAI, and Palm-Cortex-001~\cite{anil2023palm} from Google. Although the strategy to train these models has been discussed~\cite{anil2023palm, Neelakantan2022TextAC}, their detailed parameters are not known to the public, together with their capability on node classification tasks. (4) \textbf{Fine-tuned PLMs}: We consider fine-tuning Deberta on the downstream dataset, and also adopt the last hidden states of PLMs as the text embeddings.  For fine-tuning, we consider two integrating structures below. 

    \item \textit{Integration structures:} We consider \textbf{cascading structure} and \textbf{iterative structure}. (1) \textbf{Cascading structure:} we first fine-tune the PLMs on the downstream dataset. Subsequently, the text embeddings engendered by the fine-tuned PLM are employed as the initial node features for GNNs. (2) \textbf{Iterative structure:}  PLMs and GNNs are first trained separately and  further co-trained in an iterative manner by generating pseudo labels for each other. This grants us the flexibility to choose either the final iteration of PLMs or GNNs as the predictive models, which are denoted as ``GLEM-LM'' and ``GLEM-GNN'', respectively.
\end{compactenum}

We also consider non-contextualized shallow embeddings~\cite{miaschi-dellorletta-2020-contextual} including TF-IDF and Word2vec~\cite{hu2020open} as a comparison. TF-IDF is adopted to process the original text attributes for \pubmed~\cite{Sen_Namata_Bilgic_Getoor_Galligher_Eliassi-Rad_2008}, and Word2vec is utilized to encode the original text attributes for \arxiv~\cite{hu2020open}. For \arxiv and \products, we also consider the GIANT features~\cite{GIANT}, which can not be directly applied to \cora and \pubmed because of its special pre-training strategy. Furthermore, we don't include LLaMA for \arxiv and \products because it imposes an excessive computational burden when dealing with large-scale datasets.

The results are shown in Table~\ref{exp:small1}, Table~\ref{exp:small2}, and Table~\ref{exp: ogb}. In these tables, we demonstrate the performance of different combinations of text encoders and GNNs. We also include the performance of MLPs which can suggest the original quality of the textual embeddings before the aggregation. Moreover, We use colors to show the top 3 best LLMs under each GNN (or MLP) model. Specifically, We use \textcolor{yellow}{yellow} to denote the best one under a specific GNN/MLP model, \textcolor{green}{green} the second best one, and \textcolor{pink}{pink} the third best one.

\begin{table*}[!ht]
\caption{Experimental results for feature-level \textit{LLMs-as-Enhancer} on \cora and \pubmed with a low labeling ratio. Since MLPs do not provide structural information, it is meaningless to co-train it with PLM (with their performance shown as N/A). We use \textcolor{yellow}{yellow} to denote the best performance under a specific GNN/MLP model, \textcolor{green}{green} the second best one, and \textcolor{pink}{pink} the third best one.}
\label{exp:small1}
\centering
\resizebox{0.75\linewidth}{!}{
\begin{tabular}{lcccccc}
\toprule
& \multicolumn{3}{c}{\textsc{cora}} & \multicolumn{3}{c}{\textsc{pubmed}} \\ 
\cmidrule(lr){2-4}\cmidrule(lr){5-7}
& GCN & GAT & MLP & GCN & GAT & MLP \\ 
\midrule
\multicolumn{7}{l}{\textbf{Non-contextualized Shallow Embeddings}} \\ 
TF-IDF & 81.99 ± 0.63 & 82.30 ± 0.65 & 67.18 ± 1.01 & 78.86 ± 2.00 & 77.65 ± 0.91 & 71.07 ± 0.78 \\ 
Word2Vec & 74.01 ± 1.24 & 72.32 ± 0.17 & 55.34 ± 1.31 & 70.10 ± 1.80 & 69.30 ± 0.66 & 63.48 ± 0.54 \\ 
\midrule
\multicolumn{7}{l}{\textbf{PLM/LLM Embeddings without Fine-tuning}} \\ 
Deberta-base & 48.49 ± 1.86 & 51.02 ± 1.22 & 30.40 ± 0.57 & 62.08 ± 0.06 & 62.63 ± 0.27 & 53.50 ± 0.43 \\ 
LLama 7B & 66.80 ± 2.20 & 59.74 ± 1.53 & 52.88 ± 1.96 & 73.53 ± 0.06 & 67.52 ± 0.07 & 66.07 ± 0.56 \\ 
\midrule
\multicolumn{7}{l}{\textbf{Local Sentence Embedding Models}} \\ 
Sentence-BERT(MiniLM) & \cellcolor{pink}82.20 ± 0.49 & \cellcolor{green}82.77 ± 0.59 & \cellcolor{green}74.26 ± 1.44 & \cellcolor{green}81.01 ± 1.32 & 79.08 ± 0.07 & 76.66 ± 0.50 \\ 
e5-large & \cellcolor{green}82.56 ± 0.73 & 81.62 ± 1.09 & \cellcolor{yellow}74.26 ± 0.93 & \cellcolor{yellow}82.63 ± 1.13 & \cellcolor{pink}79.67 ± 0.80 & \cellcolor{yellow}80.38 ± 1.94 \\ 
\midrule
\multicolumn{7}{l}{\textbf{Online Sentence Embedding Models}} \\ 
text-ada-embedding-002 & \cellcolor{yellow}82.72 ± 0.69 & \cellcolor{pink}82.51 ± 0.86 & \cellcolor{pink}73.15 ± 0.89 & 79.09 ± 1.51 & \cellcolor{green}80.27 ± 0.41 & \cellcolor{pink}78.03 ± 1.02 \\ 
Google Palm Cortex 001 & 81.15 ± 1.01 & \cellcolor{yellow}82.79 ± 0.41 & 69.51 ± 0.83 & \cellcolor{pink}80.91 ± 0.19 & \cellcolor{yellow}80.72 ± 0.33 & \cellcolor{green}78.93 ± 0.90 \\ 
\midrule
\multicolumn{7}{l}{\textbf{Fine-tuned PLM Embeddings}} \\ 
Fine-tuned Deberta-base & 59.23 ± 1.16 & 57.38 ± 2.01 & 30.98 ± 0.68 & 62.12 ± 0.07 & 61.57 ± 0.07 & 53.65 ± 0.26 \\ 
\midrule
\multicolumn{7}{l}{\textbf{Iterative Structure}} \\ 
GLEM-GNN & 48.49 ± 1.86 & 51.02 ± 1.22 & N/A & 62.08 ± 0.06 & 62.63 ± 0.27 & N/A \\ 
GLEM-LM & 59.23 ± 1.16 & 57.38 ± 2.01 & N/A & 62.12 ± 0.07 & 61.57 ± 0.07 & N/A \\ 
\bottomrule
\end{tabular}}
\end{table*}

\begin{table*}[!ht]
\caption{Experimental results for feature-level \textit{LLMs-as-Enhancers} on \cora and \pubmed with a high labeling ratio.  We use \textcolor{yellow}{yellow} to denote the best performance under a specific GNN/MLP model, \textcolor{green}{green} the second best one, and \textcolor{pink}{pink} the third best one.}
\label{exp:small2}
\centering
\resizebox{0.75\linewidth}{!}{

\begin{tabular}{lcccccc}
\toprule
& \multicolumn{3}{c}{\textsc{cora}} & \multicolumn{3}{c}{\textsc{pubmed}} \\ 
\cmidrule(lr){2-4}\cmidrule(lr){5-7}
& GCN & GAT & MLP & GCN & GAT & MLP \\ 
\midrule
\multicolumn{7}{l}{\textbf{Non-contextualized Shallow Embeddings}} \\ 
TF-IDF & \cellcolor{yellow}90.90 ± 2.74 & \cellcolor{green}90.64 ± 3.08 & 83.98 ± 5.91 & 89.16 ± 1.25 & 89.00 ± 1.67 & 89.72 ± 3.57 \\ 
Word2Vec & 88.40 ± 2.25 & 87.62 ± 3.83 & 78.71 ± 6.32 & 85.50 ± 0.77 & 85.63 ± 0.93 & 83.80 ± 1.33 \\ 
\midrule
\multicolumn{7}{l}{\textbf{PLM/LLM Embeddings without Fine-tuning}} \\ 
Deberta-base & 65.86 ± 1.96 & 79.67 ± 3.19 & 45.64 ± 4.41 & 67.33 ± 0.69 & 67.81 ± 1.05 & 65.07 ± 0.57 \\ 
LLama 7B & 89.69 ± 1.86 & 87.66 ± 4.84 & 80.66 ± 7.72 & 88.26 ± 0.78 & 88.31 ± 2.01 & 89.39 ± 1.09 \\ 
\midrule
\multicolumn{7}{l}{\textbf{Local Sentence Embedding Models}} \\ 
Sentence-BERT(MiniLM) & 89.61 ± 3.23 & \cellcolor{yellow}90.68 ± 2.22 & \cellcolor{yellow}86.45 ± 5.56 & 90.32 ± 0.91 & 90.80 ± 2.02 & 90.59 ± 1.23 \\ 
e5-large & \cellcolor{green}90.53 ± 2.33 & 89.10 ± 3.22 & \cellcolor{green}86.19 ± 4.38 & 89.65 ± 0.85 & 89.55 ± 1.16 & 91.39 ± 0.47 \\ 
\midrule
\multicolumn{7}{l}{\textbf{Online Sentence Embedding Models}} \\ 
text-ada-embedding-002 & 89.13 ± 2.00 & \cellcolor{pink}90.42 ± 2.50 & \cellcolor{pink}85.97 ± 5.58 & 89.81 ± 0.85 & \cellcolor{pink}91.48 ± 1.94 & \cellcolor{green}92.63 ± 1.14 \\ 
Google Palm Cortex 001 & \cellcolor{pink}90.02 ± 1.86 & 90.31 ± 2.82 & 81.03 ± 2.60 & 89.78 ± 0.95 & 90.52 ± 1.35 & \cellcolor{pink}91.87 ± 0.84 \\ 
\midrule
\multicolumn{7}{l}{\textbf{Fine-tuned PLM Embeddings}} \\
Fine-tuned Deberta-base & 85.86 ± 2.28 & 86.52 ± 1.87 & 78.20 ± 2.25 & \cellcolor{pink}91.49 ± 1.92 & 89.88 ± 4.63 & \cellcolor{yellow}94.65 ± 0.13 \\ 
\midrule
\multicolumn{7}{l}{\textbf{Iterative Structure}} \\ 
GLEM-GNN & 89.13 ± 0.73 & 88.95 ± 0.64 & N/A & \cellcolor{green}92.57 ± 0.25 & \cellcolor{green}92.78 ± 0.21 & N/A \\ 
GLEM-LM & 82.71 ± 1.08 & 83.54 ± 0.99 & N/A & \cellcolor{yellow}94.36 ± 0.21 & \cellcolor{yellow}94.62 ± 0.14 & N/A \\ 
\bottomrule
\end{tabular}}
\end{table*}


\begin{table*}[ht]
\caption{Experimental results for feature-level \textit{LLMs-as-Enhancers} on \arxiv and \products dataset. MLPs do not provide structural information so it's meaningless to co-train it with PLM, thus we don't show the performance. We use \textcolor{yellow}{yellow} to denote the best performance under a specific GNN/MLP model, \textcolor{green}{green} the second best one, and \textcolor{pink}{pink} the third best one.}
\label{exp: ogb}
\centering
\resizebox{0.75\linewidth}{!}{
\begin{tabular}{lcccccc}
\toprule
& \multicolumn{3}{c}{\textsc{\arxiv}} & \multicolumn{3}{c}{\textsc{\products}} \\ \cmidrule(lr){2-4}\cmidrule(lr){5-7}
& GCN & MLP & RevGAT & SAGE & SAGN & MLP \\ \midrule
\multicolumn{7}{l}{\textbf{Non-contextualized Shallow Embeddings}} \\
TF-IDF & 72.23 ± 0.21 & 66.60 ± 0.25 & 75.16 ± 0.14 & 79.73 ± 0.48 & 84.40 ± 0.07 & 64.42 ± 0.18 \\ 
Word2Vec & 71.74 ± 0.29 & 55.50 ± 0.23 & 73.78 ± 0.19 & 81.33 ± 0.79 & 84.12 ± 0.18 & 69.27 ± 0.54 \\ \midrule
\multicolumn{7}{l}{\textbf{PLM/LLM Embeddings without Fine-tuning}} \\
Deberta-base & 45.70 ± 5.59 & 40.33 ± 4.53 & 71.20 ± 0.48 & 62.03 ± 8.82 & 74.90 ± 0.48 & 7.18 ± 1.09 \\ \midrule
\multicolumn{7}{l}{\textbf{Local Sentence Embedding Models}} \\
Sentence-BERT(MiniLM) & 73.10 ± 0.25 & 71.62 ± 0.10 & \cellcolor{green}76.94 ± 0.11 & 82.51 ± 0.53 & 84.79 ± 0.23 & 72.73 ± 0.34 \\ 
e5-large & 73.74 ± 0.12 & \cellcolor{pink}72.75 ± 0.00 & 76.59 ± 0.44 & 82.46 ± 0.91 & \cellcolor{pink}85.47 ± 0.21 & \cellcolor{pink}77.49 ± 0.29 \\ \midrule
\multicolumn{7}{l}{\textbf{Online Sentence Embedding Models}} \\
text-ada-embedding-002 & 72.76 ± 0.23 & 72.17 ± 0.00 & \cellcolor{pink}76.64 ± 0.20 & \cellcolor{pink}82.90 ± 0.42 & 85.20 ± 0.19 & 76.42 ± 0.31 \\ \midrule
\multicolumn{7}{l}{\textbf{Fine-tuned PLM Embeddings}} \\
Fine-tuned Deberta-base & \cellcolor{pink}74.65 ± 0.12 & \cellcolor{green}72.90 ± 0.11 & 75.80 ± 0.39 & 82.15 ± 0.16 & 84.01 ± 0.05 & \cellcolor{green}79.08 ± 0.23 \\ \midrule
\textbf{Others} \\
GIANT & 73.29 ± 0.10 & \cellcolor{yellow}73.06 ± 0.11 & 75.90 ± 0.19 & \cellcolor{green}83.16 ± 0.19 & \cellcolor{green}86.67 ± 0.09 & \cellcolor{yellow}79.82 ± 0.07 \\ \midrule
\multicolumn{7}{l}{\textbf{Iterative Structure}} \\ 
GLEM-GNN & \cellcolor{yellow}75.93 ± 0.19 & N/A & \cellcolor{yellow}76.97 ± 0.19 & \cellcolor{yellow}83.16 ± 0.09 & \cellcolor{yellow}87.36 ± 0.07 & N/A \\ 
GLEM-LM & \cellcolor{green}75.71 ± 0.24 & N/A & 75.45 ± 0.12 & 81.25 ± 0.15 & 84.83 ± 0.04 & N/A \\ 
\bottomrule
\end{tabular}}
\end{table*}

\subsubsection{Node Classification Performance Comparison}

\vspace{1em}
\textbf{\uline{Observation 1.} Combined with different types of text embeddings, GNNs demonstrate distinct effectiveness.}

From Table~\ref{exp: ogb}, if we compare the performance of TF-IDF and fine-tuned PLM embeddings when MLP is the predictor, we can see that the latter usually achieves much better performance. However, when a GNN model is adopted as the predictor, the performance of TF-IDF embedding is close to and even surpasses the PLM embedding. This result is consistent with the findings in~\cite{Purchase2022RevisitingEF}, which suggests that GNNs present distinct effectiveness for different types of text embeddings. However, we don't find a simple metric to determine the effectiveness of GNNs on different text embeddings. We will further discuss this limitation in Section~\ref{sec: limit}.

\textbf{\uline{Observation 2.} Fine-tune-based LLMs may fail at low labeling rate settings.}

From Table~\ref{exp:small1}, we note that no matter the cascading structure or the iterative structure, fine-tune-based LLMs' embeddings perform poorly for low labeling rate settings. Both fine-tuned PLM and GLEM present a large gap against deep sentence embedding models and TF-IDF, which do not involve fine-tuning. When training samples are limited, fine-tuning may fail to transfer sufficient knowledge for the downstream tasks.

\textbf{\uline{Observation 3.} With a simple cascading structure, the combination of deep sentence embedding with GNNs makes a strong baseline.}

From Table~\ref{exp:small1}, Table~\ref{exp:small2}, Table~\ref{exp: ogb}, we can see that with a simple cascading structure, the combination of deep sentence embedding models (including both local sentence embedding models and online sentence embedding models) with GNNs show competitive performance,  under all dataset split settings. The intriguing aspect is that, during the pre-training stage of these deep sentence embedding models, no structural information is incorporated. Therefore, it is astonishing that these structure-unaware models can outperform GIANT on \arxiv, which entails a structure-aware self-supervised learning stage.

\textbf{\uline{Observation 4.} Simply enlarging the model size of LLMs may not help with the node classification performance.}

From Table~\ref{exp:small1} and Table~\ref{exp:small2}, we can see that although the performance of the embeddings generated by LLaMA outperforms the Deberta-base without fine-tuning by a large margin, there is still a large performance gap between the performance of embeddings generated by deep sentence embedding models in the low labeling rate setting. This result indicates that simply increasing the model size may not be sufficient to generate high-quality embeddings for node classification. The pre-training objective may be an important factor.

\subsubsection{Scalability Investigation} 
\vspace{1em}
In the aforementioned experimental process, we empirically find that in larger datasets like \arxiv, methods like GLEM that require fine-tuning of the PLMs will take several orders of magnitude more time in the training stage than these that do not require fine-tuning. It presents a hurdle for these approaches to be applied to even larger datasets or scenarios with limited computing resources. To gain a more comprehensive understanding of the efficiency and scalability of different LLMs and integrating structures, we conduct an experiment to measure the running time and memory usage of different approaches. It should be noted that we mainly consider the scalability problem in the training stage,  which is different from the efficiency problem in the inference stage.

In this study, we choose representative models from each type of LLMs, and each kind of integrating structure.  For TF-IDF, it's a shallow embedding that doesn't involve either training or inference, so the time and memory complexity of the LM phase can be neglected. In terms of Sentence-BERT, for the LM phase, this kind of local sentence embedding model does not involve a fine-tuning stage, and they only need to generate the initial embeddings. For text-ada-embedding-002, which is offered as an API service, we make API calls to generate embeddings. In this part, we set the batch size of Ada to 1,024 and call the API asynchronously, then we measure the time consumption to generate embeddings as the LM phase running time. For Deberta-base, we record the time used to fine-tune the model and generate the text embeddings as the LM phase running time. For GLEM, since it co-trains the PLM and GNNs, we consider LM phase running time and GNN phase running time together (and show the total training time in the ``LM phase'' column). The efficiency results are shown in Table~\ref{exp: eff}. We also report the peak memory usage in the table. We adopt the default output dimension of each text encoder, which is shown in the brackets. 

\begin{table*}[!ht]
\caption{Efficiency analysis on \arxiv. Note that we show the dimension of generated embeddings in the brackets. For GIANT, it adopts a special pre-training stage, which will introduce computation overhead with orders of magnitude larger than that of fine-tuning. The specific time was not discussed in the original paper, therefore its cost in LM-phase is not shown in the table. } 
\label{exp: eff}
\centering
 \resizebox{0.85\linewidth}{!}{
\begin{tabular}{clcccc}
\toprule
\textbf{Input features}                       & \textbf{Backbone} & \textbf{\begin{tabular}[c]{@{}c@{}} LM-phase \\ Running time(s)\end{tabular}} & \textbf{\begin{tabular}[c]{@{}c@{}} LM-phase \\ Memory (GB)\end{tabular}} & \textbf{\begin{tabular}[c]{@{}c@{}} GNN-phase \\ Running time(s)\end{tabular}} & \textbf{\begin{tabular}[c]{@{}c@{}} GNN-phase \\ Memory (GB)\end{tabular}} \\ \midrule
\multirow{2}{*}{\begin{tabular}[c]{@{}c@{}} \textbf{TF-IDF} \\ {(1024)}\end{tabular}}              & GCN               & N/A                                 & N/A                            & 53                                 & 9.81                          \\ 
                                              & RevGAT            & N/A                                 & N/A                            & 873                                & 7.32                          \\ \midrule
\multirow{2}{*}{{\begin{tabular}[c]{@{}c@{}} \textbf{Sentence-BERT} \\ {(384)}\end{tabular}}}               & GCN               & 239                               & 1.30                         & 48                                 & 7.11                          \\ 
                                              & RevGAT            & 239                               & 1.30                         & 674                                & 4.37                          \\ \midrule
\multirow{2}{*}{\begin{tabular}[c]{@{}c@{}} \textbf{text-ada-embedding-002} \\ {(1536)}\end{tabular}}                 & GCN               & 165                               & N/A                            & 73                                 & 11.00                         \\ 
                                              & RevGAT            & 165                               & N/A                            & 1038                               & 8.33                          \\ \midrule
\multirow{2}{*}{\begin{tabular}[c]{@{}c@{}} \textbf{Deberta-base} \\ {(768)}\end{tabular}} & GCN               & 13560                             & 12.53                        & 50                                 & 9.60                          \\ 
                                              & RevGAT            & 13560                             & 12.53                        & 122                                & 6.82                          \\ \midrule
\multirow{2}{*}{{\begin{tabular}[c]{@{}c@{}} \textbf{GLEM-GNN} \\ {(768)}\end{tabular}}}            & GCN               & 68071                             & 18.22                        & N/A                              & N/A                         \\ 
                                              & RevGAT            & 68294                             & 18.22                        & N/A                              & N/A                         \\ \midrule
\multirow{2}{*}{\begin{tabular}[c]{@{}c@{}} \textbf{GIANT} \\ {(768)}\end{tabular}} & GCN               &  N/A                           & N/A                        & 50                                 & 9.60                          \\ 
                                              & RevGAT            & N/A                             & N/A                       & 122                                & 6.82                          \\ \bottomrule
\end{tabular}}
\end{table*}

\textbf{\uline{Observation 5.} For integrating structures,  iterative structure introduces massive computation overhead in the training stage.} 


From Table~\ref{exp:small2} and Table~\ref{exp: ogb}, GLEM presents a superior performance in datasets with an adequate number of labeled training samples, especially in large-scale datasets like \arxiv and \products. However, from Table~\ref{exp: eff}, we can see that it introduces massive computation overhead in the training stage compared to Deberta-base with a cascading structure, which indicates the potential efficiency problem of the iterative structures. 


Moreover, from Table~\ref{exp: eff}, we note that for the GNN phase, the dimension of initial node features, which is the default output dimension of text encoders mainly determines memory usage and time cost.  

\textbf{\uline{Observation 6.} In terms of different LLM types, deep sentence embedding models present better efficiency in the training stage.}

In Table~\ref{exp: eff}, we analyze the efficiency of different types of LLMs by selecting representative models from each category. Comparing fine-tune-based PLMs with deep sentence embedding models, we observe that the latter demonstrates significantly better time efficiency as they do not require a fine-tuning stage. Additionally, deep sentence embedding models exhibit improved memory efficiency as they solely involve the inference stage without the need to store additional information such as gradients.

\subsection{Text-level Enhancement}
\label{sec: tle}
\begin{figure}[!htb]
    \centering
    \includegraphics[width=\linewidth]{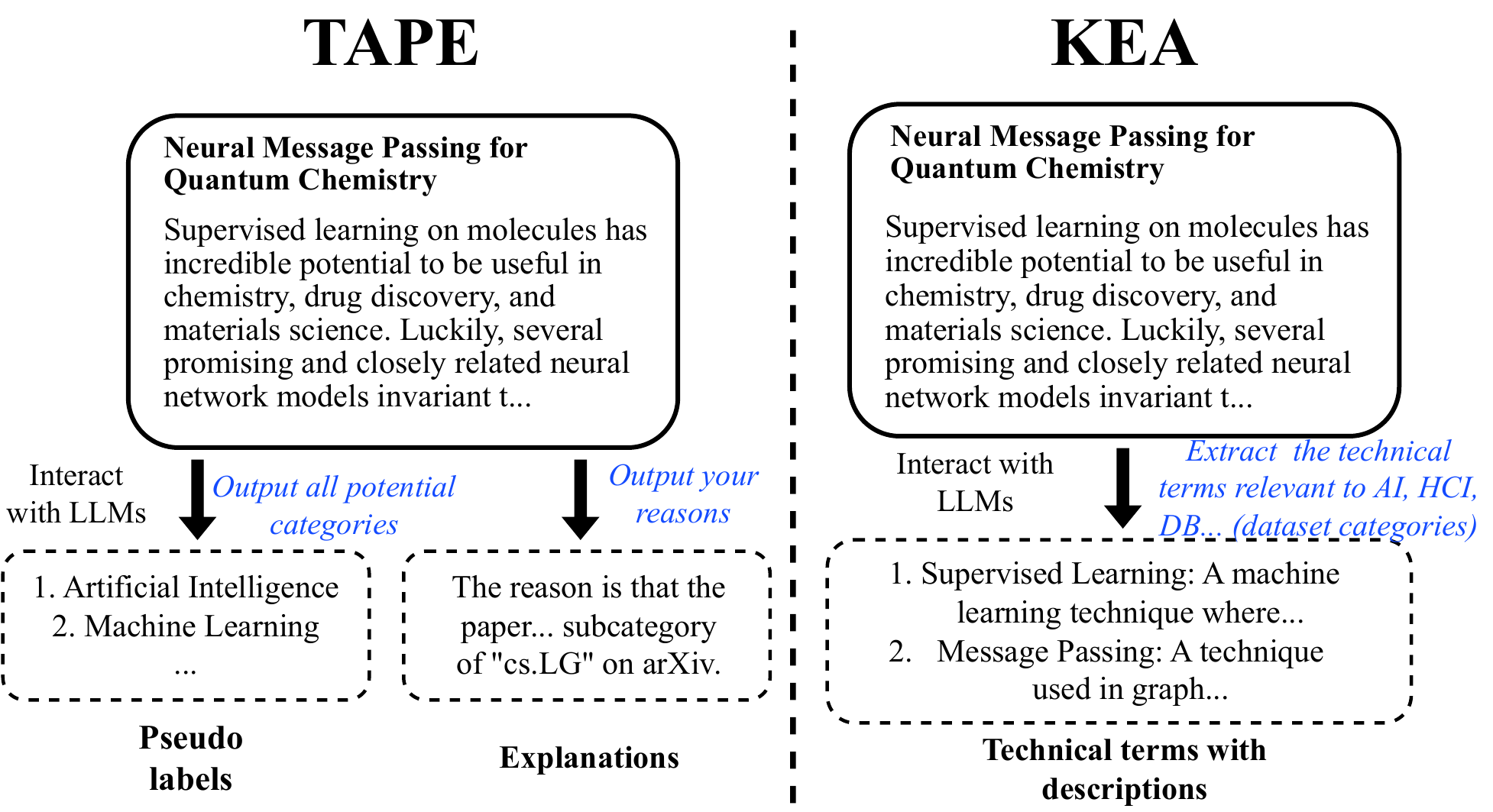}
    \caption{Illustrations for TAPE and KEA. TAPE leverages the knowledge of LLMs to generate explanations for their predictions. For KEA, we prompt the LLMs to generate a list of technical terms with their descriptions. The main motivation is to augment the attribute information.}
    \label{fig:textlevel}
\end{figure}

\vspace{1em}
For feature-level enhancement, LLMs in the pipeline must be embedding-visible. However, the most powerful LLMs such as ChatGPT~\cite{OpenAI2022}, PaLM~\cite{anil2023palm}, and GPT4~\cite{OpenAI2023GPT4TR} are all deployed as online services~\cite{sun2022black}, which put strict restrictions so that users can not get access to model parameters and embeddings. Users can only interact with these embedding-invisible LLMs through texts, which means that user inputs must be formatted as texts and LLMs will only yield text outputs.
In this section, we explore the potential for these embedding-invisible LLMs to do text-level enhancement. To enhance the text attribute at the text level, the key is to expand more information that is not contained in the original text attributes. Based on this motivation and a recent paper~\cite{he2023explanations}, we study the following two potential text-level enhancements, and illustrative examples of these two augmentations are shown in Figure~\ref{fig:textlevel}. 

\begin{compactenum}[1.] 
    \item \textbf{TAPE}~\cite{he2023explanations}: The motivation of TAPE is to leverage the knowledge of LLMs to generate high-quality node features. Specifically, it uses LLMs to generate pseudo labels and explanations. These explanations aim to make the logical relationship between the text features and corresponding labels more clear. For example, given the original attributes ``mean-field approximation" and the ground truth label ``probabilistic methods'', it will generate a description such as ``mean-field approximation is a widely adopted simplification technique for probabilistic models'', which makes the connection of these two attributes much more clear.  After generating pseudo labels and explanations, they further adopt PLMs to be fine-tuned on both the original text attributes and the explanations generated by LLMs, separately. Next, they generate the corresponding text features and augmented text features based on the original text attributes and augmented text attributes respectively, and finally ensemble them together as the initial node features for GNNs.
    \item \textbf{Knowledge-Enhanced Augmentation}: The motivation behind Knowledge-Enhanced Augmentation (KEA) is to enrich the text attributes by providing additional information. KEA is inspired by knowledge-enhanced PLMs such as ERNIE~\cite{Sun2019ERNIEER} and K-BERT~\cite{Liu2019KBERTEL} and aims to explicitly incorporate external knowledge. 
    In KEA, we prompt the LLMs to generate a list of knowledge entities along with their text descriptions. For example, we can generate a description for the abstract term ``Hopf-Rinow theorem'' as follows: ``The Hopf-Rinow theorem establishes that a Riemannian manifold, which is both complete and connected, is geodesically complete if and only if it is simply connected.'' By providing such descriptions, we establish a clearer connection between the theorem and the category ``Riemannian geometry''. Once we obtain the entity list, we encode it either together with the original text attribute or separately. We try encoding text attributes with fine-tuned PLMs and deep sentence embedding models. We also employ ensemble methods to combine these embeddings. One potential advantage of KEA is that it is loosely coupled with the prediction performance of LLMs. In cases where LLMs generate incorrect predictions, TAPE may potentially generate low-quality node features because the explanations provided by PLMs may also be incorrect. However, with KEA, the augmented features may exhibit better stability since we do not rely on explicit predictions from LLMs.
\end{compactenum}

\subsubsection{Experimental Setups} 
\vspace{1em}
To evaluate these two strategies, we conduct experiments on two small datasets \cora and \pubmed considering the cost to use the LLMs. For low labeling ratio and high labeling ratio, we adopt the same setting as that in Table~\ref{exp:small1} and Table~\ref{exp:small2}. For predictors, we adopt GCN, GAT, and MLP to study both the quality of textual embeddings before and after aggregations. For LLMs, we adopt ChatGPT with the latest version (gpt-3.5-turbo-0613). To better understand the effectiveness of TAPE, we separate it into TA, P, and E, where ``TA'' refers to ``text attributes'', ``P'' refers to ``pseudo labels'', and ``E'' refers to ``explanations''. For KEA, we try two approaches to inject the augmented textual attributes. The first approach is appending the augmented textual attributes into the original attribute, which is denoted as ``KEA-I''. Then the combined attributes are encoded into features. The second approach is to encode the augmented attributes and original attributes separately, which is denoted as ``KEA-S''. We report the results for original, augmented, and ensembling features. Both TAPE and KEA adopt the cascading structures. After encoding the text attributes with LLMs, the generated embeddings are adopted as the initial features for GNNs. We try two approaches to encode the attributes, which are fine-tuned PLMs and local sentence embedding models. Specifically, we adopt Deberta-base and e5-large. To conduct a fair comparison, we first determine the better text encoder by evaluating their overall performance. Once the text encoder is selected, we proceed to compare the performance of the augmented attributes against the original attributes.



\vspace{0.5em}
\noindent{}\textbf{A comprehensive evaluation of TAPE.} We first gain a deeper understanding of TAPE through a comprehensive ablation study. The experimental results are shown in Table~\ref{tab:TAPEabla} and Table~\ref{tab:TAPEabla2}. We show the approach we adopt to encode the text attributes in the bracket. In particular, we mainly consider fine-tuned Deberta-base, which is denoted as PLM, and e5-large, which is denoted as e5. 

\begin{table*}[!ht]
\centering
\caption{A detailed ablation study of TAPE on \cora and \pubmed dataset in low labeling rate setting. For each combination of features and models, we use \textcolor{yellow}{yellow} to denote the best performance under a specific GNN/MLP model, \textcolor{green}{green} the second best one, and \textcolor{pink}{pink} the third best one. }
\label{tab:TAPEabla}
\centering
\resizebox{0.75\linewidth}{!}{
\begin{tabular}{@{}llcccccc@{}}
\toprule
 &
   & 
  \multicolumn{3}{c}{\cora} &
  \multicolumn{3}{c}{\pubmed} \\  \cmidrule(lr){3-5} \cmidrule(lr){6-8} 
\multirow{-2}{*}{} &
  \multirow{-2}{*}{} &
  \textbf{GCN} &
  \textbf{GAT} &
  \textbf{MLP} &
  \textbf{GCN} &
  \textbf{GAT} &
  \textbf{MLP} \\ \midrule
 &
  \textbf{TAPE}  &
  74.56 ± 2.03 &
  75.27 ± 2.10 &
  64.44 ± 0.60 &
  \cellcolor{green}85.97 ± 0.31 &
  \cellcolor{green}86.97 ± 0.33 &
  \cellcolor{green}93.18 ± 0.28 \\
 &
  \textbf{P} &
  52.79 ± 1.47 &
  62.13 ± 1.50 &
  63.56 ± 0.52 &
  81.92 ± 1.89 &
  \cellcolor{yellow}88.27 ± 0.01 &
  \cellcolor{yellow}93.27 ± 0.15 \\
 &
  \textbf{TA + E} (e5) &
  \cellcolor{green}83.38 ± 0.42 &
  \cellcolor{yellow}84.00 ± 0.09 &
  \cellcolor{yellow}75.73 ± 0.53 &
  \cellcolor{yellow}87.44 ± 0.49 &
  \cellcolor{pink}86.71 ± 0.92 &
  \cellcolor{pink}90.25 ± 1.56 \\
 &
  \textbf{TA + E} (PLM) &
  78.02 ± 0.56 &
  64.08 ± 12.36 &
  55.72 ± 11.98 &
  80.70 ± 1.73 &
  79.66 ± 3.08 &
  76.42 ± 2.18 \\
 &
  \textbf{E} (PLM) &
  79.46 ± 1.10 &
  74.82 ± 1.19 &
  63.04 ± 0.88 &
  81.88 ± 0.05 &
  81.56 ± 0.07 &
  76.90 ± 1.60 \\
\multirow{-6}{*}{\begin{tabular}[c]{@{}l@{}}\uline{\textbf{TAPE}}\end{tabular}} &
  \textbf{E} (e5) &
  \cellcolor{yellow}84.38 ± 0.36 &
  \cellcolor{green}83.01 ± 0.60 &
  \cellcolor{pink}70.64 ± 1.10 &
  82.23 ± 0.78 &
  80.30 ± 0.77 &
  77.23 ± 0.48 \\ \midrule
 &
  \textbf{TA} (PLM) &
  59.23 ± 1.16 &
  57.38 ± 2.01 &
  30.98 ± 0.68 &
  62.12 ± 0.07 &
  61.57 ± 0.07 &
  53.65 ± 0.26 \\ 
\multirow{-2}{*}{\begin{tabular}[c]{@{}l@{}}\uline{\textbf{Original}} \\ \uline{\textbf{attributes}}\end{tabular}} &
  \textbf{TA} (e5) &
  \cellcolor{pink}82.56 ± 0.73 &
  \cellcolor{pink}81.62 ± 1.09 &
  \cellcolor{green}74.26 ± 0.93 &
  \cellcolor{pink}82.63 ± 1.13 &
  79.67 ± 0.80 &
  80.38 ± 1.94 \\ \bottomrule
\end{tabular}}
\end{table*}

\begin{table*}[!ht]
\centering
\caption{A detailed ablation study of TAPE on \cora and \pubmed dataset in the high labeling rate setting. For each combination of features and models, we use \textcolor{yellow}{yellow} to denote the best performance under a specific GNN/MLP model, \textcolor{green}{green} the second best one, and \textcolor{pink}{pink} the third best one. }
\label{tab:TAPEabla2}
\resizebox{0.75\linewidth}{!}{
\centering
\begin{tabular}{@{}llcccccc@{}}
\toprule
 &
   &
  \multicolumn{3}{c}{\cora} &
  \multicolumn{3}{c}{\pubmed} \\ \cmidrule(l){3-8} 
 &
   &
  \textbf{GCN} &
  \textbf{GAT} &
  \textbf{MLP} &
  \textbf{GCN} &
  \textbf{GAT} &
  \textbf{MLP} \\ \midrule
\multirow{6}{*}{{\ul \textbf{TAPE}}} &
  \textbf{TAPE} &
  87.88 ± 0.98 &
  88.69 ± 1.13 &
  83.09 ± 0.91 &
  \cellcolor{green}92.22 ± 1.30 &
  \cellcolor{green}93.35 ± 1.50 &
  \cellcolor{green}95.05 ± 0.27 \\
 &
  \textbf{P} &
  64.90 ± 1.39 &
  80.11 ± 4.01 &
  70.31 ± 1.91 &
  85.73 ± 0.59 &
  91.60 ± 0.62 &
  93.65 ± 0.35 \\
 &
  \textbf{TA + E} (e5) &
  \cellcolor{yellow}90.68 ± 2.12 &
  \cellcolor{yellow}91.86 ± 1.36 &
  \cellcolor{yellow}87.00 ± 4.83 &
  \cellcolor{yellow}92.64 ± 1.00 &
  \cellcolor{yellow}93.35 ± 1.24 &
  94.34 ± 0.86 \\
 &
  \textbf{TA + E} (PLM) &
  87.44 ± 1.74 &
  88.40 ± 1.60 &
  82.80 ± 1.00 &
  90.23 ± 0.71 &
  \cellcolor{pink}91.73 ± 1.58 &
  \cellcolor{yellow}95.40 ± 0.32 \\
 &
  \textbf{E} (PLM) &
  83.28 ± 4.53 &
  82.47 ± 6.06 &
  80.41 ± 3.35 &
  88.90 ± 2.94 &
  83.00 ± 14.07 &
  87.75 ± 14.83 \\
 &
  \textbf{E} (e5) &
  \cellcolor{pink}89.39 ± 2.69 &
  \cellcolor{green}90.13 ± 2.52 &
  \cellcolor{pink}84.05 ± 4.03 &
  89.68 ± 0.78 &
  90.61 ± 1.61 &
  91.09 ± 0.85 \\ \midrule
\multirow{2}{*}{\textbf{\begin{tabular}[c]{@{}l@{}}\uline{Original}\\ \uline{attributes}\end{tabular}}} &
  \textbf{TA} (PLM) &
  85.86 ± 2.28 &
  86.52 ± 1.87 &
  78.20 ± 2.25 &
  \cellcolor{pink}91.49 ± 1.92 &
  89.88 ± 4.63 &
  \cellcolor{pink}94.65 ± 0.13 \\
 &
  \textbf{TA} (e5) &
  \cellcolor{green}90.53 ± 2.33 &
  \cellcolor{pink}89.10 ± 3.22 &
  \cellcolor{green}86.19 ± 4.38 &
  89.65 ± 0.85 &
  89.55 ± 1.16 &
  91.39 ± 0.47 \\ \bottomrule
\end{tabular}}
\end{table*}

\textbf{\uline{Observation 7.} The effectiveness of TAPE is mainly from the explanations \textbf{E} generated by LLMs.}  

From the ablation study, we can see that compared to pseudo labels \textbf{P}, the explanations present better stability across different datasets. One main advantage of adopting explanations generated by LLMs  is that these augmented attributes present better performance in the low-labeling rate setting. From Table~\ref{tab:TAPEabla}, we note that when choosing PLM as the encoders, \textbf{E} performs much better than \textbf{TA} in the low labeling rate setting. Compared to explanations, we find that the effectiveness of the P mainly depends on the zero-shot performance of LLMs, which may present large variances across different datasets. In the following analysis, we use \textbf{TA + E} and neglect the pseudo labels generated by LLMs.

\textbf{\uline{Observation 8.} Replacing fine-tuned PLMs with deep sentence embedding models can further improve the overall performance of TAPE.}

From Table~\ref{tab:TAPEabla} and Table~\ref{tab:TAPEabla2}, we observe that adopting e5-large as the LLMs to encode the text attributes can achieve good performance across different datasets and different data splits. Specifically, the \textbf{TA + E} encoded with e5 can achieve top 3 performance in almost all settings. In the following analysis, we adopt e5 to encode the original and enhanced attributes \textbf{TA + E}.

\paragraph{Effectiveness of \textbf{KEA}} We then show the results of \textbf{KEA} in Table~\ref{tab:kea1} and Table~\ref{tab:kea2}. For \textbf{KEA-I}, we inject the description of each technical term directly into the original attribute. For \textbf{KEA-S}, we encode the generated description and original attribute separately.

\textbf{\uline{Observation 9.} The proposed knowledge enhancement attributes \textbf{KEA} can enhance the performance of the original attribute \textbf{TA}.} 

From Table~\ref{tab:kea1} and Table~\ref{tab:kea2}, we first compare the performance of features encoded by e5 and PLM. We see that the proposed \textbf{KEA} is more fitted to the e5 encoder, and fine-tuned PLM embeddings present poor performance on the low labeling rate, thus we also select e5 as the encoder to further compare the quality of attributes. From Table~\ref{tab: kea4} we can see that the proposed \textbf{KEA-I + TA} and \textbf{KEA-S + TA} attributes can consistently outperform the original attributes \textbf{TA}. 

\textbf{\uline{Observation 10.} For different datasets, the most effective enhancement methods may vary.}

Moreover, we compare the performance of our proposed \textbf{KEA} with \textbf{TA + E}, and the results are shown in Table~\ref{tab: kea4}. We can see that on \cora, our methods can achieve better performance while \textbf{TA + E} can achieve better performance on \pubmed. One potential explanation for this phenomenon is that \textbf{TA + E} relies more on the capability of LLMs. Although we have removed the pseudo labels \textbf{P}, we find that the explanations still contain LLMs' predictions. As a result, the effectiveness of $\textbf{TA + E}$ will be influenced by LLMs' performance on the dataset. As shown in \cite{he2023explanations}, the LLMs can achieve superior performance on the \pubmed dataset but perform poorly on the \cora dataset. Compared to \textbf{TA + E}, our proposed \textbf{KEA} only utilizes the commonsense knowledge of the LLMs, which may have better stability across different datasets.



\begin{table*}[!ht]
\centering
\caption{A detailed ablation study of KEA on \cora and \pubmed dataset in the low labeling rate setting. For each combination of features and models, we use \textcolor{yellow}{yellow} to denote the best performance under a specific GNN/MLP model, \textcolor{green}{green} the second best one, and \textcolor{pink}{pink} the third best one.}
\label{tab:kea1}
\centering
\resizebox{0.75\linewidth}{!}{
\begin{tabular}{@{}llcccccc@{}}
\toprule
 &
   &
  \multicolumn{3}{c}{\cora} &
  \multicolumn{3}{c}{\pubmed} \\ \cmidrule(l){3-8} 
 &
   &
  \textbf{GCN} &
  \textbf{GAT} &
  \textbf{MLP} &
  \textbf{GCN} &
  \textbf{GAT} &
  \textbf{MLP} \\ \midrule
\multirow{2}{*}{\textbf{\begin{tabular}[c]{@{}l@{}}\uline{Original}\\ \uline{attributes}\end{tabular}}} &
  TA (PLM) &
  59.23 ± 1.16 &
  57.38 ± 2.01 &
  30.98 ± 0.68 &
  62.12 ± 0.07 &
  61.57 ± 0.07 &
  53.65 ± 0.26 \\
 &
  TA (e5) &
  82.56 ± 0.73 &
  81.62 ± 1.09 &
  \cellcolor{pink}74.26 ± 0.93 &
  \cellcolor{pink}82.63 ± 1.13 &
  79.67 ± 0.80 &
  \cellcolor{pink}80.38 ± 1.94 \\ \midrule
\multirow{8}{*}{{\ul \textbf{KEA}}} &
  KEA-I + TA (e5) &
  \cellcolor{pink}83.20 ± 0.56 &
  \cellcolor{green}83.38 ± 0.63 &
  \cellcolor{green}74.34 ± 0.97 &
  \cellcolor{yellow}83.30 ± 1.75 &
  \cellcolor{green}81.16 ± 0.87 &
  \cellcolor{green}80.74 ± 2.44 \\
 &
  KEA-I + TA (PLM) &
  53.21 ± 11.54 &
  55.38 ± 4.64 &
  31.80 ± 3.63 &
  57.13 ± 8.20 &
  58.66 ± 4.27 &
  52.28 ± 4.47 \\
 &
  KEA-I (e5) &
  81.35 ± 0.77 &
  82.04 ± 0.72 &
  70.64 ± 1.10 &
  81.98 ± 0.91 &
  \cellcolor{pink}81.04 ± 1.39 &
  79.73 ± 1.63 \\
 &
  KEA-I (PLM) &
  36.68 ± 18.63 &
  37.69 ± 12.79 &
  30.46 ± 0.60 &
  56.22 ± 7.17 &
  59.33 ± 1.69 &
  52.79 ± 0.51 \\
 &
  KEA-S + TA (e5) &
  \cellcolor{yellow}84.63 ± 0.58 &
  \cellcolor{yellow}85.02 ± 0.40 &
  \cellcolor{yellow}76.11 ± 2.66 &
  \cellcolor{green}82.93 ± 2.38 &
  \cellcolor{yellow}81.34 ± 1.51 &
  \cellcolor{yellow}80.74 ± 2.44 \\
 &
  KEA-S + TA (PLM) &
  51.36 ± 16.13 &
  52.85 ± 7.00 &
  34.56 ± 5.09 &
  59.47 ± 6.09 &
  51.93 ± 3.27 &
  51.11 ± 2.63 \\
 &
  KEA-S (e5) &
  \cellcolor{green}84.38 ± 0.36 &
  \cellcolor{pink}83.01 ± 0.60 &
  70.64 ± 1.10 &
  82.23 ± 0.78 &
  80.30 ± 0.77 &
  77.23 ± 0.48 \\
 &
  KEA-S (PLM) &
  28.97 ± 18.24 &
  43.88 ± 10.31 &
  30.36 ± 0.58 &
  61.22 ± 0.94 &
  54.93 ± 1.55 &
  47.94 ± 0.89 \\ \bottomrule
\end{tabular}}
\end{table*}

\begin{table*}[!ht]
\centering
\caption{A detailed ablation study of KEA on \cora and \pubmed dataset in the high labeling rate setting. For each combination of features and models, we use \textcolor{yellow}{yellow} to denote the best performance under a specific GNN/MLP model, \textcolor{green}{green} the second best one, and \textcolor{pink}{pink} the third best one.}
\label{tab:kea2}
\centering
\resizebox{0.75\linewidth}{!}{
\begin{tabular}{@{}llcccccc@{}}
\toprule
 &
   &
  \multicolumn{3}{c}{\cora} &
  \multicolumn{3}{c}{\pubmed} \\ \cmidrule(l){3-8} 
 &
   &
  \textbf{GCN} &
  \textbf{GAT} &
  \textbf{MLP} &
  \textbf{GCN} &
  \textbf{GAT} &
  \textbf{MLP} \\ \midrule
\multirow{2}{*}{\begin{tabular}[c]{@{}l@{}}\uline{\textbf{Original}}\\ \uline{\textbf{Attributes}}\end{tabular}} &
  \textbf{TA} (PLM) &
  85.86 ± 2.28 &
  86.52 ± 1.87 &
  78.20 ± 2.25 &
  \cellcolor{pink}91.49 ± 1.92 &
  89.88 ± 4.63 &
  \cellcolor{pink}94.65 ± 0.13 \\
 &
  \textbf{TA} (e5) &
  90.53 ± 2.33 &
  89.10 ± 3.22 &
  86.19 ± 4.38 &
  89.65 ± 0.85 &
  89.55 ± 1.16 &
  91.39 ± 0.47 \\ \midrule
\multirow{8}{*}{\uline{\textbf{KEA}}} &
\textbf{KEA-I + TA} (e5) &
  \cellcolor{yellow}91.12 ± 1.76 &
  \cellcolor{green}90.24 ± 2.93 &
  \cellcolor{green}87.88 ± 4.44 &
  90.19 ± 0.83 &
  90.60 ± 1.22 &
  92.12 ± 0.74 \\
 &
 \textbf{KEA-I + TA} (PLM) &
  87.07 ± 1.04 &
  87.66 ± 0.86 &
  79.12 ± 2.77 &
  \cellcolor{yellow}92.32 ± 0.64 &
  \cellcolor{yellow}92.29 ± 1.43 &
  \cellcolor{yellow}94.85 ± 0.20 \\
 &
  \textbf{KEA-I} (e5) &
  \cellcolor{green}91.09 ± 1.78 &
  90.13 ± 2.76 &
  \cellcolor{pink}86.78 ± 4.12 &
  89.56 ± 0.82 &
  90.25 ± 1.34 &
  91.92 ± 0.80 \\
 &
 \textbf{KEA-I} (PLM) &
  86.08 ± 2.35 &
  85.23 ± 3.15 &
  77.97 ± 2.87 &
  \cellcolor{green}91.73 ± 0.58 &
  \cellcolor{green}91.93 ± 1.76 &
  \cellcolor{green}94.76 ± 0.33 \\
 &
  
  \textbf{KEA-S + TA} (e5) &
  \cellcolor{pink}91.09 ± 1.78 &
  \cellcolor{yellow}92.30 ± 1.69 &
  \cellcolor{yellow}88.95 ± 4.96 &
  90.40 ± 0.92 &
  \cellcolor{pink}90.82 ± 1.30 &
  91.78 ± 0.56 \\
 &
  \textbf{KEA-S + TA} (PLM) &
  83.98 ± 5.13 &
  87.33 ± 1.68 &
  80.04 ± 1.32 &
  86.11 ± 5.68 &
  89.04 ± 5.82 &
  94.35 ± 0.48 \\ &
  \textbf{KEA-S} (e5) &
  89.39 ± 2.69 &
  \cellcolor{pink}90.13 ± 2.52 &
  84.05 ± 4.03 &
  89.68 ± 0.78 &
  90.61 ± 1.61 &
  91.09 ± 0.85 \\
 &
  
  \textbf{KEA-S} (PLM) &
  83.35 ± 7.30 &
  85.67 ± 2.00 &
  76.76 ± 1.82 &
  79.68 ± 19.57 &
  69.90 ± 19.75 &
  85.91 ± 6.47 \\
  \bottomrule
\end{tabular}}
\end{table*}

\begin{table*}[!ht]
\centering
\caption{Comparison of the performance of TA, KEA-I, and KEA-S, and TA + E. The best performance is shown with an underline. \cora (low) means a low labeling rate setting, and \cora (high) denotes a high labeling rate setting.}
\label{tab: kea4}
\centering
\resizebox{0.75\linewidth}{!}{
\begin{tabular}{@{}lllllll@{}}
\toprule
 &
  \multicolumn{3}{c}{\cora (low)} &
  \multicolumn{3}{c}{\pubmed (low)} \\ \cmidrule(l){2-7} 
 &
  \multicolumn{1}{c}{\textbf{GCN}} &
  \multicolumn{1}{c}{\textbf{GAT}} &
  \multicolumn{1}{c}{\textbf{MLP}} &
  \multicolumn{1}{c}{\textbf{GCN}} &
  \multicolumn{1}{c}{\textbf{GAT}} &
  \multicolumn{1}{c}{\textbf{MLP}} \\ \midrule
\textbf{TA} &
  82.56 ± 0.73 &
  81.62 ± 1.09 &
  74.26 ± 0.93 &
  82.63 ± 1.13 &
  79.67 ± 0.80 &
  80.38 ± 1.94 \\
\textbf{KEA-I + TA} &
  83.20 ± 0.56 &
  83.38 ± 0.63 &
  74.34 ± 0.97 &
  83.30 ± 1.75 &
  81.16 ± 0.87 &
  80.74 ± 2.44 \\
\textbf{KEA-S + TA} &
  {\ul 84.63 ± 0.58} &
  {\ul 85.02 ± 0.40} &
  {\ul 76.11 ± 2.66} &
  82.93 ± 2.38 &
  81.34 ± 1.51 &
  80.74 ± 2.44 \\
\textbf{TA+E} &
  83.38 ± 0.42 &
  84.00 ± 0.09 &
  75.73 ± 0.53 &
  {\ul 87.44 ± 0.49} &
  {\ul 86.71 ± 0.92} &
  {\ul 90.25 ± 1.56} \\ \midrule
 &
  \multicolumn{3}{c}{\cora (high)} &
  \multicolumn{3}{c}{\pubmed (high)} \\ \cmidrule(l){2-7} 
 &
  \multicolumn{1}{c}{\textbf{GCN}} &
  \multicolumn{1}{c}{\textbf{GAT}} &
  \multicolumn{1}{c}{\textbf{MLP}} &
  \multicolumn{1}{c}{\textbf{GCN}} &
  \multicolumn{1}{c}{\textbf{GAT}} &
  \multicolumn{1}{c}{\textbf{MLP}} \\ \midrule
\textbf{TA} &
  90.53 ± 2.33 &
  89.10 ± 3.22 &
  86.19 ± 4.38 &
  89.65 ± 0.85 &
  89.55 ± 1.16 &
  91.39 ± 0.47 \\
\textbf{KEA-I + TA} &
  {\ul 91.12 ± 1.76} &
  90.24 ± 2.93 &
  87.88 ± 4.44 &
  90.19 ± 0.83 &
  90.60 ± 1.22 &
  92.12 ± 0.74 \\
\textbf{KEA-S + TA} &
  91.09 ± 1.78 &
  {\ul 92.30 ± 1.69} &
  {\ul 88.95 ± 4.96} &
  90.40 ± 0.92 &
  90.82 ± 1.30 &
  91.78 ± 0.56 \\
\textbf{TA+E} &
  90.68 ± 2.12 &
  91.86 ± 1.36 &
  87.00 ± 4.83 &
  {\ul 92.64 ± 1.00} &
  {\ul 93.35 ± 1.24} &
  {\ul 94.34 ± 0.86} \\ \bottomrule
\end{tabular}}
\end{table*}

\section{LLMs as the Predictors} 
\label{sec:pred}

\vspace{1em}
In the \textit{LLMs-as-Enhancers} pipeline, the role of the LLMs remains somewhat limited since we only utilize their pre-trained knowledge but overlook their reasoning capability. Drawing inspiration from the LLMs' proficiency in handling complex tasks with implicit structures, such as logical reasoning~\cite{creswell2023selectioninference} and recommendation~\cite{Gao2023ChatRECTI}, we question: \textbf{Is it possible for the LLM to independently perform predictive tasks on graph structures?} By shifting our focus to node attributes and overlooking the graph structures, we can perceive node classification as a text classification problem. In~\cite{Sun2023TextCV}, the LLMs demonstrate significant promise, suggesting that they can proficiently process text attributes. However, one key problem is that LLMs are not originally designed to process graph structures. Therefore, it can not directly process structural information like GNNs. 

In this section, we aim to explore the potential of LLMs as a predictor. In particular, we first check whether LLM can perform well without any structural information. Then, we further explore some prompts to incorporate structural information with natural languages. Finally, we show a case study in Section~\ref{sec: anno} to explore its potential usage as an annotator for graphs.

\subsection{How Can LLM Perform on Popular Graph Benchmarks without Structural Information?}
\label{sec: tc}
\vspace{1em}
In this subsection, we treat the node classification problem as a text classification problem by ignoring the structural information. We adopt ChatGPT (gpt-3.5-turbo-0613) as the LLMs to conduct all the experiments. 
We choose five popular textual graph datasets with raw text attributes: \cora~\cite{McCallum2000AutomatingTC}, \citeseer~\cite{giles1998citeseer}, \pubmed~\cite{Sen_Namata_Bilgic_Getoor_Galligher_Eliassi-Rad_2008}, \arxiv, and \products~\cite{hu2020open}. The details of these datasets can be found in Appendix~\ref{app: real-world}. Considering the costs to query LLMs' APIs, it's not possible for us to test the whole dataset for these graphs. Considering the rate limit imposed by OpenAI\footnote{\url{https://platform.openai.com/docs/guides/rate-limits/overview}}, we randomly select 200 nodes from the test sets as our test data. In order to ensure that these 200 nodes better represent the performance of the entire set, we repeat all experiments twice. Additionally, we employ zero-shot performance as a sanity check, comparing it with the results in TAPE~\cite{he2023explanations} to ensure minimal discrepancies.

We explore the following strategies:

\begin{compactenum}[1.]
    \item \textbf{Zero-shot prompts}: This approach solely involves the attribute of a given node.
    \item \textbf{Few-shot prompts}: On the basis of zero-shot prompts, few-shot prompts provide in-context learning samples together with their labels for LLMs to better understand the task. In addition to the node's content, this approach integrates the content and labels of randomly selected in-context samples from the training set. In the section, we adopt random sampling to select few-shot prompts. 
    \item \textbf{Zero-shot prompts with Chain-of-Thoughts (CoT)}: CoT~\cite{wei2022chain} presents its effectiveness in various reasoning tasks, which can greatly improve LLMs' reasoning abilities. In this study, we test whether CoT can improve LLMs' capability on node classification tasks. On the basis of zero-shot prompts, we guide the LLMs to generate the thought process by using the prompt "think it step by step".
    \item \textbf{Few-shot prompts with CoT}: Inspired by~\cite{Zhang2022AutomaticCO}, which demonstrates that incorporating the CoT process generated by LLMs can further improve LLMs' reasoning capabilities. Building upon the few-shot prompts, this approach enables the LLMs to generate a step-by-step thought process for the in-context samples. Subsequently, the generated CoT processes are inserted into the prompt as auxiliary information. 
\end{compactenum}

\noindent{}\textbf{Output Parsing.}
In addition, we need a parser to extract the output from LLMs. 
We devise a straightforward approach to retrieve the predictions from the outputs. 
Initially, we instruct the LLMs to generate the results in a formatted output like ``a python list''.  
Then, we can use the symbols ``['' and ``]'' to locate the expected outputs. 
It should be noted that this design aims to extract the information more easily but has little influence on the performance. 
We observe that sometimes LLMs will output contents that are slightly different from the expected format, for example, output the expected format ``Information Retrieval'' to ``Information Extraction''. 
In such cases, we compute the edit distance between the extracted output and the category names and select the one with the smallest distance. 
This method proves effective when the input context is relatively short. 
If this strategy encounters errors, we resort to extracting the first mentioned categories in the output texts as the predictions. 
If there's no match, then the model's prediction for the node is incorrect.

To reduce the variance of LLMs' predictions, we set the temperature to 0. 
For few-shot cases, we find that providing too much context will cause LLMs to generate outputs that are not compatible with the expected formats. 
Therefore, we set a maximum number of samples to ensure that LLMs generate outputs with valid formats. 
In this study, we choose this number to $2$ and adopt accuracy as the performance metric.

\begin{table*}[!ht]
\footnotesize
\caption{Performance of LLMs on real-world text attributed graphs without structural information, we also include the result of GCN (or SAGE for \products) together with Sentence-BERT features. For \cora, \citeseer, \pubmed, we show the results of the low labeling rate setting. }
\label{tab: llmres1}
    \centering
    \resizebox{0.75\linewidth}{!}{
    \centering
    \begin{tabular}{cccccc}
    \toprule
        \textbf{} & \textsc{\cora} & \textsc{\citeseer} & \textsc{\pubmed} & \textsc{\arxiv} & \textsc{\products} \\ \midrule
        \textbf{Zero-shot} & 67.00 ± 1.41 & 65.50 ± 3.53   & 90.75 ± 5.30 & 51.75 ± 3.89  & 70.75 ± 2.48 \\ 
        \textbf{Few-shot} & 67.75 ± 3.53 & 66.00 ± 5.66 & 85.50 ± 2.80  & 50.25 ± 1.06 & 77.75 ± 1.06 \\ 
        \textbf{Zero-shot with COT} & 64.00 ± 0.71 & 66.50 ± 2.82  & 86.25 ± 3.29  & 50.50 ± 1.41 & 71.25 ± 1.06 \\ 
        \textbf{Few-shot with COT} & 64.00 ± 1.41  & 60.50 ± 4.94  & 85.50 ± 4.94 & 47.25 ± 2.47 & 73.25 ± 1.77 \\ 
        \textbf{GCN/SAGE} & 82.20 ± 0.49  & 71.19 ± 1.10 & 81.01 ± 1.32 & 73.10 ± 0.25 &  82.51 ± 0.53\\ \bottomrule
    \end{tabular}}
\end{table*}

\subsubsection{Observations}
\vspace{1em}

\textbf{\uline{Observation 11.} LLMs present preliminary effectiveness on some datasets.}


According to the results in Table~\ref{tab: llmres1}, it is evident that LLMs demonstrate remarkable zero-shot performance on \pubmed. When it comes to \products, LLMs can achieve performance levels comparable to fine-tuned PLMs. However, there is a noticeable performance gap between LLMs and GNNs on \cora and \pubmed datasets. To gain a deeper understanding of this observation, it is essential to analyze the output of LLMs.

\textbf{\uline{Observation 12.} Wrong predictions made by LLMs are sometimes also reasonable.}

After investigating the output of LLMs, we find that a part of the wrong predictions made by LLMs are very reasonable. An example is shown in Table~\ref{table:reason}. In this example, we can see that besides the ground truth label "Reinforcement Learning", "Neural Networks" is also a reasonable label, which also appears in the texts. We find that this is a common problem for \cora, \citeseer, and \arxiv. For \arxiv, there are usually multiple labels for one paper on the website. However, in the \arxiv dataset, only one of them is chosen as the ground truth. This leads to a misalignment between LLMs' commonsense knowledge and the annotation bias inherent in these datasets. Moreover, we find that introducing few-shot samples presents little help to mitigate the annotation bias. 

\begin{table}[h!]
\caption {A wrong but reasonable prediction made by LLMs }
\label{table:reason}
\centering
\rule{\linewidth}{2pt}
\parbox{\linewidth}{
\vspace{5pt}
         \textbf{Paper:} The Neural Network House: An overview; Typical home comfort systems utilize only rudimentary forms of energy management and conservation. The most sophisticated technology in common use today is an automatic setback thermostat. Tremendous potential remains for improving the efficiency of electric and gas usage... \newline
    \textbf{Ground Truth: } Reinforcement Learning \newline
    \textbf{LLM's Prediction: } Neural Networks \newline
\vspace{2pt}
}
\rule{\linewidth}{2pt}
\vspace{5pt}
\end{table}

\textbf{\uline{Observation 13.} Chain-of-thoughts do not bring in performance gain.}

For reasoning tasks in the general domain, chain-of-thoughts is believed to be an effective approach to increase LLM's reasoning capability~\cite{wei2022chain}. However, we find that it's not effective for the node classification task. This phenomenon can be potentially explained by \textbf{Observation 12}. In contrast to mathematical reasoning, where a single answer is typically expected, multiple reasonable chains of thought can exist for node classification. An example is shown in Table~\ref{table:cot_wrong}. This phenomenon poses a challenge for LLMs as they may struggle to match the ground truth labels due to the presence of multiple reasonable labels.

\begin{table}[h!]
\caption {An example that LLMs generate CoT processes not matching with ground truth labels}
\label{table:cot_wrong}
\centering
\rule{\linewidth}{2pt}
\parbox{\linewidth}{
\vspace{5pt}
         \textbf{Paper:} The Neural Network House: An overview.: Typical home comfort systems utilize only rudimentary forms of energy management and conservation. The most sophisticated technology in common use today is an automatic setback thermostat. Tremendous potential remains for improving the efficiency of electric and gas usage... \newline
    \textbf{Generated Chain-of-thoughts:} The paper discusses the use of neural networks for intelligent control and mentions the utilization of neural network reinforcement learning and prediction techniques. Therefore, the most likely category for this paper is 'Neural Networks'. \newline
    \textbf{Ground Truth: } Reinforcement Learning \newline
    \textbf{LLM's Prediction: } Neural Networks \newline
\vspace{2pt}
}
\rule{\linewidth}{2pt}
\vspace{5pt}
\end{table}



\textbf{\uline{Observation 14.} For prompts that are very similar in semantics, there may be huge differences in their effects.}

In addition, we observe that TAPE \cite{he2023explanations} implements a unique prompt on the \arxiv dataset, yielding impressive results via zero-shot prompts. The primary distinction between their prompts and ours lies in the label design. Given that all papers originate from the computer science subcategory of Arxiv, they employ the brief term "arxiv cs subcategories" as a substitute for these 40 categories. Remarkably, this minor alteration contributes to a substantial enhancement in performance. To delve deeper into this phenomenon, we experiment with three disparate label designs: (1) Strategy 1: the original Arxiv identifier, such as "arxiv cs.CV"; (2) Strategy 2: natural language descriptors, like "computer vision"; and (3) Strategy 3: the specialized prompt, utilizing "arxiv cs subcategory" to denote all categories. Unexpectedly, we discover that Strategy 3 significantly outperforms the other two (refer to Table~\ref{tab: arxiv}). 

\begin{table}[!ht]
\footnotesize
\centering
\caption{Performance of LLMs on OGB-Arxiv dataset, with three different label designs. }
\label{tab: arxiv}
\begin{tabular}{cccc}
\toprule
      & \textbf{Strategy 1} & \textbf{Strategy 2} & \textbf{Strategy 3} \\ \cmidrule{2-4}
\textsc{\arxiv} & 48.5       &   51.8    &  74.5        \\ \bottomrule
\end{tabular}
\end{table}

Given that LLMs undergo pre-training on extensive text corpora, it's likely that these corpora include papers from the Arxiv database. That specific prompt could potentially enhance the ``activation'' of these models' corresponding memory. However, the reason for the excellent results achieved by this kind of prompt might not stem from the simple data memorization of the LLM~\cite{huang2023can}. When applying to papers after 2023 that are not included in the pre-training corpus of the LLMs, this prompt also achieves similar effectiveness. This phenomenon reminds us that when using ChatGPT, sometimes providing more information in the prompt (such as category information from the \arxiv dataset) may actually lead to a decrease in performance.


\subsection{Incorporating Structural Information in the Prompts}
\label{sec: saware}
\vspace{1.5em}
As we note, LLMs can already present superior zero-shot performance on some datasets without providing any structural information. However, there is still a large performance gap between LLMs and GNNs in \cora, \citeseer, and \arxiv. 
Then a question naturally raises that \textit{whether we can further increase LLMs' performance by incorporating structural information?}
To answer this problem, we first need to identify how to denote the structural information in the prompt. LLMs such as ChatGPT are not originally designed for graph structures, so they can not process adjacency matrices like GNNs. In this part, we study several ways to convey structural information and test their effectiveness on the \cora dataset. 

Specifically, we first consider inputting the whole graph into the LLMs. 
Using \cora dataset as an example, we try to use prompts like ``node 1: \prompt{paper content}'' to represent attributes, and prompts like ``node 1 cites node 2'' to represent the edge. 
However, we find that this approach is not feasible since LLMs usually present a small input context length restriction. 
As a result, we consider an ``ego-graph'' view, which refers to the subgraphs induced from the center nodes. In this way, we can narrow the number of nodes to be considered. 

Specifically, we first organize the neighbors of the current nodes as a list of dictionaries consisting of attributes and labels of the neighboring nodes for training nodes. Then, the LLMs summarize the neighborhood information. It should be noted that we only consider 2-hop neighbors because GNNs typically have 2 layers, indicating that the 2-hop neighbor information is the most useful in most cases. Considering the input context limit of LLMs, we empirically find that each time we can summarize the attribute information of 5 neighbors. In this paper, we sample neighbors once and only summarize those selected neighbors. In practice, we can sample multiple times and summarize each of them to obtain more fine-grained neighborhood information. 

\begin{table*}[!ht]
\footnotesize
\caption{Performance of LLMs on real-world text attributed graphs with summarized neighborhood information. For \cora, \citeseer, \pubmed, we show the results of the low labeling rate setting. We also include the result of GCN (or SAGE for \products) together with Sentence-BERT features.}
\label{tab: llmres2}
    \centering
    \resizebox{0.75\linewidth}{!}{
    \begin{tabular}{cccccc}
    \toprule
        \textbf{} & \textsc{\cora} & \textsc{\citeseer} & \textsc{\pubmed} & \textsc{\arxiv} & \textsc{\products} \\ \midrule
        \textbf{Zero-shot} & 67.00 ± 1.41 & 65.50 ± 3.53   & 90.75 ± 5.30 & 51.75 ± 3.89  & 70.75 ± 2.48 \\ 
        \textbf{Few-shot} & 67.75 ± 3.53 & 66.00 ± 5.66 & 85.50 ± 2.80  & 50.25 ± 1.06 & 77.75 ± 1.06 \\ 
        \textbf{Zero-Shot with 2-hop info} & 71.75 ± 0.35 & 62.00 ± 1.41 & 88.00 ± 1.41 & 55.00 ± 2.83 & 75.25 ± 3.53 \\ 
        \textbf{Few-Shot with 2-hop info} & 74.00 ± 4.24 & 67.00 ± 4.94 & 79.25 ± 6.71 & 52.25 ± 3.18 & 76.00 ± 2.82  \\ 
        \textbf{GCN/SAGE} & 82.20 ± 0.49  & 71.19 ± 1.10 & 81.01 ± 1.32 & 73.10 ± 0.25 &  82.51 ± 0.53\\ \bottomrule
    \end{tabular}}
\end{table*}

\textbf{\uline{Observation 15.} Neighborhood summarization is likely to achieve performance gain.}

From Table~\ref{tab: llmres2}, we note that incorporating neighborhood information in either zero-shot or few-shot approaches yields performance gains compared to the zero-shot prompt without structural information except on the \pubmed dataset. By following the "homophily" assumption~\cite{homo, mao2023demystifying}, which suggests that neighboring nodes tend to share the same labels, the inclusion of neighboring information can potentially alleviate annotation bias. For instance, let's consider a paper from Arxiv covering general topics like transformers. Merely analyzing the content of this paper makes it difficult to determine which category the author would choose, as categories such as "Artificial Intelligence," "Machine Learning," and "Computer Vision" are all plausible options. However, by examining its citation relationships, we can better infer the author's bias. If the paper cites numerous sources from the "Computer Vision" domain, it is likely that the author is also a researcher in that field, thereby favoring the selection of this category. Consequently, structural information provides implicit supervision to assist LLMs in capturing the inherent annotation bias in the dataset. However, from the \pubmed dataset, we observe that incorporating neighborhood information results in clear performance drop, which necessitates a deep analysis below.

\textbf{\uline{Observation 16.} LLMs with structure prompts may suffer from heterophilous neighboring nodes.}

From Table~\ref{tab: llmres2}, we observe that LLMs perform worse on \pubmed after incorporating the structural information. To gain a deeper understanding, we focus on those nodes where zero-shot prompts without structural information can lead to correct prediction but prompts with 2-hop information can't.

An example of this kind of node is shown in Table~\ref{tab:pubcase1}. 
After analyzing the 2-hop neighbors of this node, we find that 15 out of 19 2-hop neighboring nodes have different labels against this node. This case is usually denoted as "heterophily"~\cite{homo}, which is a phenomenon in graph theory where nodes in a graph tend to connect with nodes that are dissimilar to them. In this case, we find that both GNNs and LLMs with a structure-aware prompt make wrong predictions. However, LLMs ignoring structural information get correct predictions, which indicates that LLMs with a structure-aware prompt may also suffer from the "heterophily" problem. 

\begin{table}[h!]
\caption {GNNs and LLMs with structure-aware prompts are both wrong}
\label{tab:pubcase1}
\centering
\rule{\linewidth}{2pt}
\parbox{\linewidth}{
\vspace{5pt}
Paper:
 Title: C-reactive protein and incident cardiovascular events among men with diabetes. \\
Abstract: OBJECTIVE: Several large prospective studies have shown that baseline levels of C-reactive protein (CRP)  ... \\ 
Neighbor Summary: 
 This paper focuses on different aspects of \textbf{type 2 diabetes} mellitus. It explores the levels of various markers such as tumor necrosis factor-alpha, interleukin-2 ... \\
 \textbf{Ground truth: "Diabetes Mellitus Type 1"} \\
 \textbf{Structure-ignorant prompts: "Diabetes Mellitus Type 1"} \\
 \textbf{Structure-aware prompt: "Diabetes Mellitus Type 2"} \\
 \textbf{GNN: "Diabetes Mellitus Type 2"} \\
\vspace{2pt}
}
\rule{\linewidth}{2pt}
\vspace{5pt}
\end{table}

\subsection{Case Study: LLMs as the Pseudo Annotators}
\label{sec: anno}
\vspace{1.5em}

From Table~\ref{tab: llmres1}, we show that LLMs can be good \textbf{zero-shot predictors} on several real-world graphs, which provides the possibility to conduct zero-shot inference on datasets without labels. Despite the effectiveness of LLMs, it still presents two problems: (1) The price of using LLMs' API is not cheap, and conducting inference on all testing nodes for large graphs incurs high costs; (2) Whether it is a locally deployed open-source LLM or a closed source LLM accessed through an API, the inference with these LLMs are much slower than GNNs, since the former has high computational resource requirements, while the latter has rate limits. One potential solution to these challenges is leveraging the knowledge of LLMs to train smaller models like GNNs, which inspires a potential application of LLMs to be used as annotators. 

Based on the preliminary experimental outcomes, LLMs display encouraging results on certain datasets, thus highlighting their potential for generating high-quality pseudo-labels. However, the use of LLMs as an annotator introduces a new challenge. A key consideration lies in deciding the nodes that should be annotated. Unlike the self-labeling in GNNs\cite{noisy, DRGST, li2023informative}, where confidence-based or information-based metrics are employed to estimate the quality of pseudo-labels. It remains a difficult task to determine the confidence of pseudo-labels generated by LLMs. Additionally, different nodes within a graph have distinct impacts on other nodes \cite{Wu2019ActiveLF}. Annotating certain nodes can result in a more significant performance improvement compared to others. Consequently, the primary challenge can be summarized as follows: how can we effectively select both the critical nodes within the graph and the reliable nodes in the context of LLMs?

Taking into account the complexity of these two challenges, we don't intend to comprehensively address them in this paper. Instead, we present a preliminary study to evaluate the performance of a simple strategy: randomly selecting a subset of nodes for annotation. It is worth noting  that advanced selection strategies such  as active learning~\cite{Wu2019ActiveLF} could be adopted to improve the final performance. We leave such exploration as future work.
Regarding the annotation budget, we adopt a "low labeling rate" setting, wherein we randomly select a total of 20 nodes multiplied by the number of classes. For the selected nodes, we adopt 75\% of them as training nodes and the rest as validation nodes. Consequently, we annotate a total of 140 nodes in the \cora dataset and 60 nodes in the \pubmed dataset. In this part, we use GCN as the GNN model and adopt the embeddings generated by the Sentence-BERT model. The results are shown in Table~\ref{tab:anno}. We can observe that training GCN on the pseudo labels can lead to satisfying performance. Particularly, it can match the performance of GCN trained on ground truth labels with 10 shots per class.
As a reference, around 67\% of the pseudo labels for \cora can match ground truth labels, while around 93\% of the pseudo labels for \pubmed are ground truth labels.

\begin{table}[!ht]
\centering
\caption{Performance of GCN trained on either pseudo labels generated by LLMs, or ground truth labels}
\begin{tabular}{lcc}
\toprule
                    & \textbf{\cora} & \textbf{\pubmed} \\ \hline
                    \multicolumn{3}{l}{\textit{Using pseudo labels}}                \\ 
\textbf{20 shots $\times$ \#class} & 64.95 ± 0.98  & 71.70 ± 1.06    \\ 
\multicolumn{3}{l}{\textit{Using ground truth}}                \\ 
\textbf{3 shots per class}   & 52.63 ± 1.46         & 59.35 ± 2.67    \\ 
\textbf{5 shots per class}   & 58.97 ± 1.41         & 65.98 ± 0.74           \\ 
\textbf{10 shots per class}  & 69.87 ± 2.27         & 71.51 ± 0.77           \\ \bottomrule
\end{tabular}
\label{tab:anno}
\end{table}

\textbf{\uline{Observation 17.} The quality of pseudo labels is key to downstream performance.}

Although we don't place significant emphasis on the selection of nodes to be labeled, the preliminary results show that there is relatively little variance among different random selections. Comparing this to the impact of pseudo labels, we observe that the quality of pseudo labels can make a significant difference. When higher quality pseudo labels are used, GNNs perform much better on \pubmed compared to \cora. This result highlights the importance of developing an approach to select confident nodes for LLMs.

\textbf{\uline{Observation 18.} Getting the confidence by simply prompting the LLMs may not work since they are too ``confident".}

Based on previous observations, we check some simple strategies to achieve the confidence level of LLMs' outputs.  Initially, we attempt to prompt the LLMs directly for their confidence level. However, we discover that most of the time, LLMs simply output a value of $1$, rendering it meaningless. Examples are shown in Table~\ref{table:conftry}.

\begin{table}[h!]
\caption{Prompts used to generate neighbor summary}
\label{table:conftry}
\centering
\rule{\linewidth}{2pt}
\parbox{\linewidth}{
\vspace{5pt}
\textbf{Instruction} \\
    Output the confidence level in the range of 0 to 1 and  the most 1 possible category of this paper as a python dict, like {''prediction": "XX", "confidence": "XX"}
\vspace{2pt}
}
\rule{\linewidth}{2pt}
\vspace{5pt}
\end{table}

Another potential solution is to utilize LLMs that support prediction logits, such as text-davinci-003. However, we observe that the probability of the outputs from these models is consistently close to 1, rendering the output not helpful.

\subsection{Case Study: Applying LLMs to handle out-of-distribution data} 
\vspace{1.5em}

Out-of-distribution (OOD) learning addresses scenarios where training and test data are drawn from different distributions. Given the ubiquity of distribution shifts in graph data~\cite{li2022out}, OOD generalization on graphs has emerged as a crucial research direction in recent years. A recent benchmark, GOOD~\cite{gui2022good}, reveals that existing GNN-based models struggle with robustness when confronted with distributional shifts. In contrast, LLMs have demonstrated commendable robustness on textual data in the presence of OOD scenarios~\cite{wang2023robustness}. Node classification on the TAG, when disregarding graph structures, can also be considered as a text classification task. Therefore, in this section, we initiate a preliminary exploration into the application of LLMs for OOD scenarios on graphs.

\textbf{Experimental Setups.} We adopt the GOOD-Arxiv dataset from the GOOD benchmark~\cite{gui2022good} considering its text attribute availability. Specifically, we adopt all four types of the OOD shift: ``Concept-degree'', ``Covariate-degree'', ``Concept-time'', and ``Covariate-time'' from the GOOD. The final results are shown in Table~\ref{tab:ood}. We adopt the prompt from TAPE~\cite{he2023explanations} since it achieves better performance on the \arxiv dataset. For comparison, we take the best baseline models from the GOOD benchmark.

\begin{table}[!ht]
\centering
\caption{OOD performance comparison. ``Val'' means the results on the IID validation sets. ``Test'' indicates the results of the OOD test sets. We can see that LLMs-as-Predictors consistently outperform the best GNN-based OOD baselines. Moreover, the gap between IID performance and OOD performance is small.}
\label{tab:ood}
\begin{tabular}{@{}lccc@{}}
\toprule
                          & Val   & Test  & Best baseline (test) \\ \midrule
\textbf{concept degree}   & 73.01 & 72.79 & 63.00                \\
\textbf{covariate degree} & 70.23 & 68.21 & 59.08                \\
\textbf{concept time}     & 72.66 & 71.98 & 67.45                \\
\textbf{covariate time}   & 74.28 & 74.37 & 71.34                \\ \bottomrule
\end{tabular}
\end{table}

\textbf{\uline{Observation 19.} LLMs-as-Predictors demonstrate robustness when facing OOD data.}

From Table~\ref{tab:ood}, we find that LLMs-as-Predictors present promising robustness against OOD data. It should be noted that we only try a simple structure-ignorant prompt, and we may further improve the OOD performance of LLMs by selecting proper in-context samples and incorporating structural information. In a nutshell, LLMs present great potential to enhance the OOD generalization capability of graph models. 

\section{Related Work}
\label{sec:rw}

\vspace{1em}
Following our proposed two pipelines, i.e., LLMs as the Enhancers and LLMs as the Predictors, we review existing works in this section.

\subsection{LLMs as the Enhancers}
\vspace{1em}
In the recent surge of research, increasing attention has been paid on the intersection of LLMs and GNNs in the realm of TAGs~\cite{GLEM, GIANT, yasunaga2022linkbert, yasunaga2022dragon, Purchase2022RevisitingEF, he2023explanations, Zhu2021TextGNNIT, Hu2020GPTGNNGP, liu2023one, duan2023simteg}. 
 Compared to shallow embeddings, LLMs can provide a richer repository of commonsense knowledge, which could potentially enhance the performance of downstream tasks~\cite{Qiu2020PretrainedMF}.   
 
Several studies employ PLMs as text encoders, transforming text attributes into node features, which can thus be classified as \fen. The integration structures vary among these works: some adopt a simple cascading structure~\cite{Purchase2022RevisitingEF, GIANT, yasunaga2022linkbert, liu-etal-2020-fine}, while others opt for an iterative structure~\cite{GLEM, graphformers, yasunaga2022dragon}. 
For those utilizing the cascading structure, preliminary investigations have been conducted to determine how the quality of text embeddings affects downstream classification performance~\cite{Purchase2022RevisitingEF}. 
GIANT~\cite{GIANT} attempts to incorporate structural information into the pre-training stage of PLMs, achieving improved performance albeit with additional training overhead. 
SimTEG~\cite{duan2023simteg} suggests that using embeddings obtained through efficiently fine-tuned parameters to replace the original embeddings of pre-trained language models can solve the problem of overfitting during fine-tuning, thereby further enhancing the performance of the cascading structure. 
OneForAll~\cite{liu2023one} further adopts sentence embedding model to unify the feature space, and propose a unified model for diverse tasks across multiple datasets. 
This cascading structure has also been successfully applied to tasks such as fact verification~\cite{liu-etal-2020-fine} and question answering~\cite{yasunaga2022linkbert}. 
However, despite its simplicity, recent studies~\cite{GLEM} have identified potential drawbacks of the cascading structure. Specifically, it establishes a tenuous connection between the text attribute and the graph. The embeddings generated by the PLMs do not take graph structures into account, and the parameters of the PLMs remain constant during the GNN training process. 
Alternatively, in the iterative structure, Graphformers~\cite{graphformers} facilitates the co-training of PLMs and GNNs using each other's generated embeddings. GLEM~\cite{GLEM} takes this a step further by considering pseudo labels generated by both PLMs and GNNs and incorporating them into the optimization process. DRAGON~\cite{yasunaga2022dragon} successfully extends the iterative structure to the knowledge graph domain.

Compared to these studies focusing on PLMs, a recent stud\-y~\cite{he2023explanations} considers the usage of embedding-invisible LLMs such as ChatGPT~\cite{OpenAI2022} for representation learning on TAGs, which aims to adopt LLMs to enhance the text attributes and thus can be categorized into \ten. 
This work introduces a prompt designed to generate explanations for the predictions made by LLMs. These generated explanations are subsequently encoded into augmented features by PLMs. Through the ensemble of these augmented features with the original features, the proposed methodology demonstrates its efficacy and accomplishes state-of-the-art performance on the \arxiv leaderboard~\cite{hu2020open}. Nevertheless, the study offers limited analytical insights into the underlying reasons for the success of this approach. Additionally, we have identified a potential concern regarding the prompts utilized in the referenced study.

Another work pertaining to the integration of LLMs and GNNs is the Graph-Toolformer~\cite{graph_toolformer}. Drawing inspirations from Toolformer~\cite{schick2023toolformer}, this study utilizes LLMs as an interface to bridge the natural language commands and GNNs. This approach doesn't change the features and training of GNNs, which is out of our scope.

\subsection{LLMs as the Predictors}
\vspace{1em}

While \textit{LLMs-as-Enhancers} have proven to be effective, the pipeline still requires GNNs for final predictions. 
In a significant shift from this approach, recent studies~\cite{guo2023gpt4graph, graph_natural_language} have begun exploring a unique pipeline that solely relies on LLMs for final predictions. 
These works fall under the category of \textit{LLMs-as-Predictors}. 
The first series of work focus on applying closed-source LLMs without tuning the parameters. 
GPT4Graph~\cite{guo2023gpt4graph} evaluates the potential of LLMs in executing knowledge graph (KG) reasoning and node classification tasks. Their findings indicate that these models can deliver competitive results for short-range KG reasoning but struggle with long-range KG reasoning and node classification tasks. However, its presentation is pretty vague and they don't give the detailed format of the prompt they use. Considering the publicity of the Arxiv data, the data leakage problem in evaluation is further studied in~\cite{huang2023can}. 
NLGraph~\cite{graph_natural_language} introduces a synthetic benchmark to assess graph structure reasoning capabilities. The study primarily concentrates on traditional graph reasoning tasks such as shortest path, maximum flow, and bipartite matching, while only offering limited analysis on node classification tasks. This does not align with our central focus, primarily on graph learning, with a specific emphasis on node classification tasks.
GraphText~\cite{zhao2023graphtext} further tries to apply LLMs to a broader range of non-text-attributed graphs by converting the original features into clustering centers or pseudo labels.
LLM4Dyg~\cite{zhang2023llm4dyg} further evaluates LLMs' capability for temporal graph-related tasks. LLMGNN~\cite{chen2023label} and GPT4GNAS~\cite{wang2023graph} apply LLMs-as-predictors as annotators and agents for neural architecture search, respectively.

As these closed-source LLMs only accept text-type inputs, the first type of methods requires transforming graphs into certain form of natural language, either directly using node attributes or describing the graph structure using natural language.  
Meanwhile, due to the input length limitations of LLMs, this transformation process often results in the loss of a considerable amount of information from the graph. 
Therefore, the second type of work involves fine-tuning LLMs to enable them to understand graph information represented as embeddings. 
InstructGLM~\cite{ye2023natural} combines textual instructions with node features in embedding form, enabling LLMs to understand node features through instruction tuning. 
Subsequently, it predicts the type of nodes based on the given instructions. GraphGPT~\cite{tang2023graphgpt} further introduces cross-modal contrastive learning to align the graph and text feature spaces. 
It also introduces dual-stage instruction tuning, where the first stage adopts self-supervised instruction tuning to make LLMs better understand graph-structured information. 
The second stage adopts task-specific fine-tuning to allow LLMs achieve task-specific knowledge and then make predictions. 
GraphLLM~\cite{chai2023graphllm} and DGTL~\cite{qin2023disentangled} apply this pipeline to graph reasoning tasks and graph representation learning.

\section{Conclusions, Limitations, and Future Directions}
\label{sec:fut}
\vspace{1em}

In this section, we summarize our key findings, present the limitations of this study and discuss the potential directions of leveraging LLMs in graph machine learning. 

\subsection{Key Findings} 
\vspace{1em}
In this paper, we propose two potential pipelines: \textit{LLMs-as-Enhancers} and \textit{LLMs-as-Predictors} that incorporate LLMs to handle the text-attributed graphs. Our rigorous empirical studies reveal several interesting findings which provide new insights for future studies. We highlight some key findings below and more can be found from Observation 1 to Observation 19.

\textbf{\uline{Finding 1.} For \textit{LLMs-as-Enhancers}, deep sentence embedding models present effectiveness in terms of performance and efficiency.} We empirically find that when we adopt deep sentence embedding models as enhancers at the feature level, they present good performance under different dataset split settings, and also scalability. This indicates that they are good candidates to enhance text attributes at the feature level. 


\textbf{\uline{Finding 2.}  For \textit{LLMs-as-Enhancers}, the combination of LLMs' augmentations and ensembling demonst\-rates its effectiveness.} As demonstrated in Section~\ref{sec: tle}, when LLMs are utilized as enhancers at the text level, we observe performance improvements by ensembling the augmented attributes with the original attributes across datasets and data splits. This suggests a promising approach to enhance the performance of attribute-related tasks. The proposed pipeline involves augmenting the attributes with LLMs and subsequently ensembling the original attributes with the augmented ones. 

\textbf{\uline{Finding 3.} For \textit{LLMs-as-Predictors}, LLMs present preliminary effectiveness but also indicate potential evaluation problem.} In Section~\ref{sec:pred}, we conduct preliminary experiments on applying LLMs as predictors, utilizing both textual attributes and edge relationships. The results demonstrate that LLMs present effectiveness in processing textual attributes and achieving good zero-shot performance on certain datasets. Moreover, our analysis reveals two potential problems within the existing evaluation framework: (1) There are instances where LLMs' inaccurate predictions can also be considered reasonable, particularly in the case of citation datasets where multiple labels may be appropriate. (2) We find a potential test data leakage problem on \arxiv, which underscores the need for a careful reconsideration of how to appropriately evaluate the performance of LLMs on real-world datasets.


\subsection{Limitations} 
\label{sec: limit}

\vspace{1em}
\noindent{}{\textbf{A deeper understanding of the effectiveness of text embeddings.}} Despite the effectiveness of deep sentence embedding models, our understanding of why their embeddings outperform PLMs' on node classification tasks remains limited. Furthermore, we observe a performance gap between deep sentence embedding models and GLEM on the \products dataset, which may be related to the domains of the dataset. Moreover, as shown in Observation 4, GNNs demonstrate different levels of effectiveness on different text embeddings. However, we give limited explanations for this phenomenon. To gain a deeper understanding, we need to have a look at the original feature space and the feature space after aggregation. This phenomenon may potentially be related to the anistrophy in language model embeddings~\cite{ethayarajh-2019-contextual}. More in-depth analysis is required to better understand these phenomena.

\noindent{}{\textbf{Costs of LLM augmentations.}}
In the work, we study TAPE and KEA to enhance the textual attributes at the text level. Although these methods have proven to be effective, they require querying LLMs' APIs at least N times for a graph with N nodes. Given the cost associated with LLMs, this poses a significant expense when dealing with large-scale datasets. Consequently, we have not presented results for the \arxiv and \products datasets.

\noindent{}{\textbf{Text-formatted hand-crafted prompts to represent graphs.}} 
In Section~\ref{sec:pred}, we limit our study to the use of ``natural language'' prompts for graph representation. However, various other formats exist for representing graphs in natural language such as XML, YAML, GML, and more~\cite{Roughan2015UnravellingGF}. Moreover, we mainly design these prompts in a hand-crafted way, which is mainly based on trial and error. It's thus worthwhile to consider exploring more prompt formats and how to come up with automatic prompts.

\subsection{Future Directions}
\vspace{1em}
\noindent{}{\textbf{Extending the current pipelines to more tasks and more types of graphs.}} 
In this study, our primary focus is on investigating the node classification task for text-attributed graphs. Nevertheless, it remains unexplored whe\-ther these two pipelines can be extended to other graph-learning tasks or other types of graphs. Certain tasks necessitate the utilization of long-range information~\cite{dwivedi2022long}, and representing such information within LLMs' limited input context poses a significant challenge. Furthermore, we demonstrate that LLMs exhibit promising initial results in graphs containing abundant textual information, particularly in natural language. However, the exploration of their effective extension to other types of graphs with non-natural language information, such as molecular graph~\cite{Fey2019FastGR, Li2023EmpoweringMD}, still needs further exploration. 


\noindent{}{\textbf{Using LLMs more efficiently.}}
Despite the effectiveness of LLMs, the inherent operational efficiency and operational cost of these models still pose significant challenges. Taking ChatGPT, which is accessed through an API, as an example, the current billing model incurs high costs for processing large-scale graphs. As for locally deployed open-source large models, even just using them for inference requires substantial hardware resources, not to mention training the models with parameter updates. Therefore, developing more efficient strategies to utilize LLMs is currently a challenge.

\noindent{}{\textbf{Evaluating LLMs' capability for graph learning tasks.}}
In this paper, we briefly talk about the potential pitfalls of the current evaluation framework. There are mainly two problems: (1) the test data may already appear in the training corpus of LLMs, which is referred to as "contamination"~\footnote{\url{https://hitz-zentroa.github.io/lm-contamination/}} (2) the ground truth labels may present ambiguity, and the performance calculated based on them may not reflect LLMs' genuine capability. For the first problem, one possible mitigation is to use the latest dataset which is not included in the training corpus of LLMs. However, that means we need to keep collecting data and annotating them, which seems not an effective solution. For the second problem, one possible solution is to reconsider the ground truth design. For instance, for the categorization of academic papers, we may adopt a multi-label setting and select all applicable categories as the ground truth. However, for more general tasks, it remains a challenge to design more reasonable ground truths. Generally speaking, it's a valuable future direction to rethink how to properly evaluate LLMs.

\noindent{}{\textbf{Aligning the feature space of graph models and LLMs.}}
Currently, a major obstacle hindering the wider application of LLMs in the field of graph learning is the discrepancy between the feature space of LLMs and that of graphs. This discrepancy makes it difficult for LLMs to effectively understand information in the graph domain. There are mainly two approaches to address this issue in current work. The first approach is to translate the information on the graph into natural language that LLMs can understand. The second approach involves directly inputting the graph information in the form of embeddings and then using instruction tuning to enable LLMs to understand this information. However, both methods have their evident limitations. For the first method, the translation process can result in information loss, and the inherent input length limitation of LLMs also prevents users from inputting large-scale graphs. For the second method, the introduction of tuning significantly increases computational overhead. Is there a better way to align LLMs with graphs? A recent work targeting multimodality~\cite{pang2023frozen} has shown new possibilities. It demonstrates that with fixed LLM parameters, only a linear transformation layer is needed to convert information from the visual domain into content that can be effectively processed by LLMs, and such an architecture also holds great potential in the field of graph machine learning.


%
\bibliographystyle{abbrv}
\bibliography{references}  
%
%
\appendix
\section{Datasets} \label{app: real-world}

\vspace{1em}
In this work, we mainly use the following five real-world graph datasets. Their statistics are shown in Table~\ref{tab: alldata}.
\begin{table}[!ht]
    \centering
    \caption{Statistics of the graph datasets.}
    \label{tab: alldata}
    \resizebox{\linewidth}{!}{
    \begin{tabular}{lcccc}
    \toprule
         Dataset &  \#Nodes &\#Edges & Task  & Metric\\
         \midrule
         \cora~\cite{McCallum2000AutomatingTC}& 2,708 &5,429 & 7-class classif. & Accuracy\\
         \citeseer\textsuperscript{*}~\cite{giles1998citeseer} & 3,186 & 4,277 & 6-class classif.& Accuracy \\
         \pubmed~\cite{Sen_Namata_Bilgic_Getoor_Galligher_Eliassi-Rad_2008}& 19,717 & 44,338 & 3-class classif.& Accuracy
         \\
         \arxiv~\cite{hu2020open} & 169,343 & 1,166,243 & 40-class classif.& Accuracy
         \\
         \products~\cite{hu2020open} & 2,449,029 & 61,859,140 & 47-class classif. & Accuracy	\\
         \bottomrule
    \end{tabular}}
\end{table}

\subsection{Dataset Description}
\vspace{1em}
\label{app: datades}
In this part, we give a brief introduction to each graph dataset. It should be noted that it's cumbersome to get the raw text attributes for some datasets, and we will elaborate them below. The structural information and label information of these datasets can be achieved from Pyg~\footnote{\url{https://pytorch-geometric.readthedocs.io/en/latest/modules/data.html}}.  We will also release the pre-processed versions of these datasets to assist future related studies.

\noindent{}\textbf{\cora~\cite{McCallum2000AutomatingTC}} 
\cora is a paper citation dataset with the following seven categories: ['Rule Learning',
'Neural Networks', 'Case Based', 'Genetic Algorithms', 'Theory', 'Reinforcement Learning', 'Probabilistic Methods']. The raw text attributes can be obtained from \url{https://people.cs.umass.edu/~mccallum/data.html}. 

\noindent{}\textbf{\citeseer~\cite{giles1998citeseer}}
\citeseer is a paper citation dataset with the following seven categories: ["Agents", "ML", "IR", "DB", "HCI", "AI"]. Note that we find that the TAG versopm only contains the text attributes for 3186 nodes. As a result, we take the graph consisted of these 3186 nodes with 4277 edges. 

\noindent{}\textbf{\pubmed~\cite{Sen_Namata_Bilgic_Getoor_Galligher_Eliassi-Rad_2008}}
\pubmed is a paper citation dataset consisting scientific journals collected from the PubMed database with the following three categories: ['Diabetes Mellitus, Experimental',
'Diabetes Mellitus Type 1', 'Diabetes Mellitus Type 2']. 

\noindent{}\textbf{\arxiv and \products~\cite{hu2020open}}
These dataset are selected from the popular OGB benchmark~\cite{hu2020open}, and descriptions for these datasets can be found in \url{https://ogb.stanford.edu/docs/nodeprop}.

\section{Experiment Setups}
\vspace{0.5em}
\subsection{Computing Environment}
\vspace{1em}
We implement all the baseline models with PyG~\cite{Fey2019FastGR}, DGL~\cite{Wang2019DeepGL}, and transformers~\cite{Wolf2019HuggingFacesTS} modules. The experiments were conducted in a GPU server with eight NVIDIA RTX A5000 GPUs, each with 24GB VRAM. 

\subsection{Hyperparameters}
\vspace{1em}
For RevGAT, GraphSage, and SAGN models, we directly adopt the best hyperparameters from the OGB leaderboard~\footnote{\url{https://github.com/snap-stanford/ogb}}. For Deberta-base on \cora and \pubmed, we follow the hyperparameter setting of TAPE~\cite{he2023explanations}. In terms of GLEM, for the LM part, we follow the hyperparameter setting in their reporsitory~\footnote{\url{https://github.com/AndyJZhao/GLEM}}. For GCN, GAT, MLP, we use the following hyperparameter search range. 

\begin{compactenum}[(a)]
\item \textbf{Hidden dimension:} \{8, 16, 32, 64, 128, 256\}.
\item \textbf{Number of layers:} \{1, 2, 3\}
\item \textbf{Normalization:} \{None, BatchNorm\};
\item \textbf{Learning rate:} \{1e-2, 5e-2, 5e-3, 1e-3\}
\item \textbf{Weight Decay:} \{1e-5, 5e-5, 5e-4, 0\}
\item \textbf{Dropout:} \{0., 0.1, 0.5, 0.8\}
\item \textbf{Number of heads for GAT:} \{1, 4, 8\}
\end{compactenum}

\section{Demonstrations of TAPE}
\label{app: demo_tk}

\vspace{1em}

\noindent{}\textbf{Examples for \pubmed} After analyzing the \pubmed data\-set, we find an interesting phenomenon that sometimes the label of the paper just appears in the raw text attributes. An example is shown in Table~\ref{table:pubmed_example}. This property of \pubmed may be related to the superior zero-shot performance of LLMs on this dataset. This can also potentially explain why GCN and GAT are outperformed by MLP in the high labeling ratio. When the link between node attributes and node labels can be easily found and adequate to determine the categories, incorporating neighbors coming from other categories will introduce noise.  



\begin{table}[H]
\caption {An illustrative example for \pubmed}
\label{table:pubmed_example}
\centering
\rule{\linewidth}{2pt}
\parbox{\linewidth}{
\vspace{5pt}
          Title: Predictive power of sequential measures of albuminuria for progression to ESRD or death in Pima Indians with \textbf{type 2 diabetes}. \newline
          ... (content omitted here) \newline 
          \textbf{Ground truth label:} Diabetes Mellitus Type 2
\vspace{2pt}
}
\rule{\linewidth}{2pt}
\vspace{5pt}
\end{table}

\end{document}